\documentclass[10pt,conference]{IEEEtran}
\IEEEoverridecommandlockouts % 启用 \thanks，用于标注通讯作者脚注。

\usepackage{cite}
\usepackage{amsmath,amssymb,amsfonts}
\usepackage{algorithm}
\usepackage{algorithmic}
\usepackage{graphicx}
\usepackage{textcomp}
\usepackage[table,xcdraw]{xcolor}
\usepackage{booktabs}
\usepackage{multirow}
\usepackage[caption=false,font=footnotesize]{subfig}
\usepackage{longtable}
\usepackage{adjustbox} % 提供 max width 缩放：只缩不放，避免表格被无谓放大/缩小导致观感不一致
\usepackage{url}
\usepackage[hidelinks]{hyperref}
\usepackage{bibentry}
\usepackage{float} % 提供 [H] 参数强制浮动体放在当前位置
\usepackage{dblfloatfix}
\usepackage{placeins} % 提供 \FloatBarrier：阻止浮动体跨越指定位置

\makeatletter
\def\footnoterule{%
  \kern-3pt
  \hrule width 0.45\columnwidth height 0.4pt
  \kern 2.6pt
}
\makeatother
\graphicspath{{./}{./images/}}

\def\BibTeX{{\rm B\kern-.05em{\sc i\kern-.025em b}\kern-.08em
    T\kern-.1667em\lower.7ex\hbox{E}\kern-.125emX}}

\begin{document}

% Title
\title{Pre-PEFT Probing: Weight Statistics and Perturbation Robustness for Layer Selection in VLM Vision Encoders}

% Author info (camera-ready)：
% 按官方 IEEE conference 模板的标准 author block 写法组织，
% 使用 \and 让 IEEEtran 自动处理换行与布局。
\author{
\IEEEauthorblockN{Qingtao Xia}
\IEEEauthorblockA{\textit{Faculty of Computing} \\
\textit{Harbin Institute of Technology}\\
Harbin, China \\
\href{mailto:25s103308@stu.hit.edu.cn}{\nolinkurl{25s103308@stu.hit.edu.cn}}}
\and
\IEEEauthorblockN{Jiahua Bao}
\IEEEauthorblockA{\textit{Faculty of Computing} \\
\textit{Harbin Institute of Technology}\\
Harbin, China \\
\href{mailto:bjh_samsara@163.com}{\nolinkurl{bjh_samsara@163.com}}}
\and
\IEEEauthorblockN{Siyao Cheng\IEEEauthorrefmark{1}\thanks{\IEEEauthorrefmark{1} Corresponding author.}}
\IEEEauthorblockA{\textit{Faculty of Computing} \\
\textit{Harbin Institute of Technology}\\
Harbin, China \\
\href{mailto:csy@hit.edu.cn}{\nolinkurl{csy@hit.edu.cn}}}
\and
\IEEEauthorblockN{Jie Liu}
\IEEEauthorblockA{\textit{Faculty of Computing} \\
\textit{Harbin Institute of Technology}\\
Harbin, China \\
\href{mailto:jieliu@hit.edu.cn}{\nolinkurl{jieliu@hit.edu.cn}}}
}

% Make title and then input abstract from test.tex
\maketitle
\begin{abstract}
We propose a pre-fine-tuning probing method for Parameter-Efficient Fine-Tuning (PEFT) layer selection, aiming to obtain more stable and higher gains with fewer trainable parameters when adapting large vision--language models (VLMs). Unlike the common practice of applying LoRA and other adapters to all layers at once---where layer selection often relies on heuristic rules---we focus on the vision encoder and directly evaluate the ``adaptability'' of each Transformer layer. Specifically, we characterize each layer from two perspectives: (i) the statistical properties of its Q/K/V projection weights (e.g., norms and condition numbers); (ii) robustness under controlled parameter perturbations. We then systematically compare these indicators with the downstream performance gains brought by applying PEFT to a single layer. Across experiments covering seven benchmarks and five PEFT variants, we observe a consistent correlation: layers (or matrices) with larger weight norms and higher condition numbers are usually more robust to perturbations and are more likely to yield larger fine-tuning gains. These results show that distribution-statistics analysis and perturbation tests before fine-tuning can provide practical signals for adaptation-layer selection, thereby maintaining or improving performance while reducing trainable parameters.
\end{abstract}


% ================== Main body migrated from test.tex ==================
% All sections and figures after the abstract in test.tex
% ================== Main body ==================
\section{Introduction}
Vision--Language Models (VLMs) combine vision encoders and language decoders for multimodal understanding and generation, and models such as InternVL, LLaVA, and OpenFlamingo have shown strong image--text capabilities \cite{chen2024internvl,liu2024visual,awadalla2023openflamingo}. Most VLMs follow a modular design in which a pre-trained vision encoder is connected to an LLM through a lightweight connector \cite{minaee2024large, yin2023survey, zhang2024vision}.

For downstream adaptation, Parameter-Efficient Fine-Tuning (PEFT) is widely used because it updates only a small subset of parameters while often retaining competitive performance relative to full fine-tuning \cite{hu2021lora,houlsby2019parameter,zhang2023llama}. However, PEFT remains sensitive to where trainable modules are inserted, and layer selection is still largely heuristic.

Even so, PEFT still faces a key bottleneck in practice: hyperparameters and insertion positions are highly sensitive, and there is a lack of explainable and predictable selection criteria. For example, which layers LoRA should be inserted into, whether all layers need to be covered, and how to choose layers when ``only tuning part of the layers'' can often significantly affect the final performance, while the trial-and-error cost is high. Moreover, many fine-tuning methods introduce additional trainable parameters (e.g., low-rank adapters or gating vectors). With standard random initialization and early-stage updates, these added degrees of freedom perturb the target layer computation in a structured manner. If, before fine-tuning starts, we can pre-judge which layers are more ``worth fine-tuning'' by analyzing in-layer weight statistics and controlled perturbation responses, then there is an opportunity to improve both the efficiency and the effectiveness of PEFT.

\begin{figure*}[t]
\centering
\includegraphics[height=0.2\textheight]{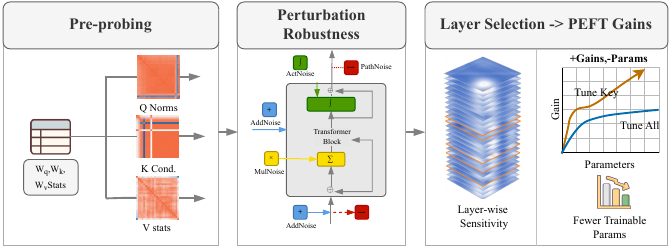}
\caption{Overview of our pre-PEFT probing pipeline for layer selection in the vision encoder. We quantify layer-wise weight statistics and perturbation robustness before fine-tuning, and use them as signals to guide PEFT layer selection.}
\label{fig:main:mainpage}
\end{figure*}

Figure~\ref{fig:main:mainpage} provides an overview of our pre-PEFT probing pipeline; perturbation details are given in Sec.~\ref{sec:perturbation_settings}.

To reduce PEFT training parameters and improve model performance, it is crucial to identify which layers will have the most significant fine-tuning effects before starting the process. In both traditional Convolutional Neural Networks (CNNs) and Vision Transformers (ViTs), deeper layers are generally more effective at learning complex features \cite{dosovitskiy2020image}. Based on this, we quantify the parameter distribution for Query (Q), Key (K), and Value (V) across each ViT layer, as well as their robustness to perturbations. We then compare the fine-tuning effectiveness of each layer using five common PEFT methods. Our empirical results indicate a stable correspondence between these pre-tuning indicators and layer-wise PEFT gains across the vision encoders of InternVL2-1B (24 layers) and Qwen2.5-VL-3B (32 layers). For each experiment, we have conducted three trials and reported the median as the result; across both backbones, run-to-run variance is small (on the order of 0.3pp). Our contributions are as follows:

\begin{itemize}
\item{We provide a unified layer-wise study linking perturbation robustness, Q/K/V weight statistics, and PEFT gains in VLM vision encoders.}
\item{Across InternVL2-1B and Qwen2.5-VL-3B, more robust layers often coincide with larger norms, higher condition numbers, and better layer-wise PEFT gains.}
\item{We show that selective tuning of a small subset of layers can remain competitive with broader tuning, providing a practical pre-screening signal for PEFT layer selection.}
\end{itemize}

The code and scripts used in this work are publicly available at \href{https://github.com/atoz03/prepeft-probing}{https://github.com/atoz03/prepeft-probing}.

\FloatBarrier
\section{Related Work}
\subsection{VLMs}
VLMs align a vision encoder with a language decoder for multimodal understanding and generation. Representative systems such as LLaVA connect CLIP ViT-L/14 to an LLM through a lightweight projector \cite{liu2024visual,radford2021learning}, while InternVL2 and Qwen2.5-VL adopt ViT-style vision encoders, lightweight connectors, and staged training recipes \cite{opengvlab2024internvl2,chen2024internvl,yang2024qwen2,bai2025qwen25vl}.

\subsection{PEFT with VLMs}
To adapt VLMs under limited budgets, PEFT updates only a small subset of parameters, with common choices including LoRA, adapters, and prompt tuning \cite{hu2021lora,houlsby2019parameter,jia2022visual}. These methods can be viewed as introducing structured perturbations to the original network during adaptation. Prior work has studied perturbation and robustness in CNN/Transformer models \cite{vaswani2017attention,coppola2024investigating,nokabadi2024reproducibility}, but their connection to layer-wise PEFT effectiveness in VLM vision encoders remains underexplored. Our work studies this connection through weight statistics, perturbation robustness, and layer-wise PEFT gains.

\FloatBarrier
\begin{figure*}[!t]
\centering
\includegraphics[width=\textwidth]{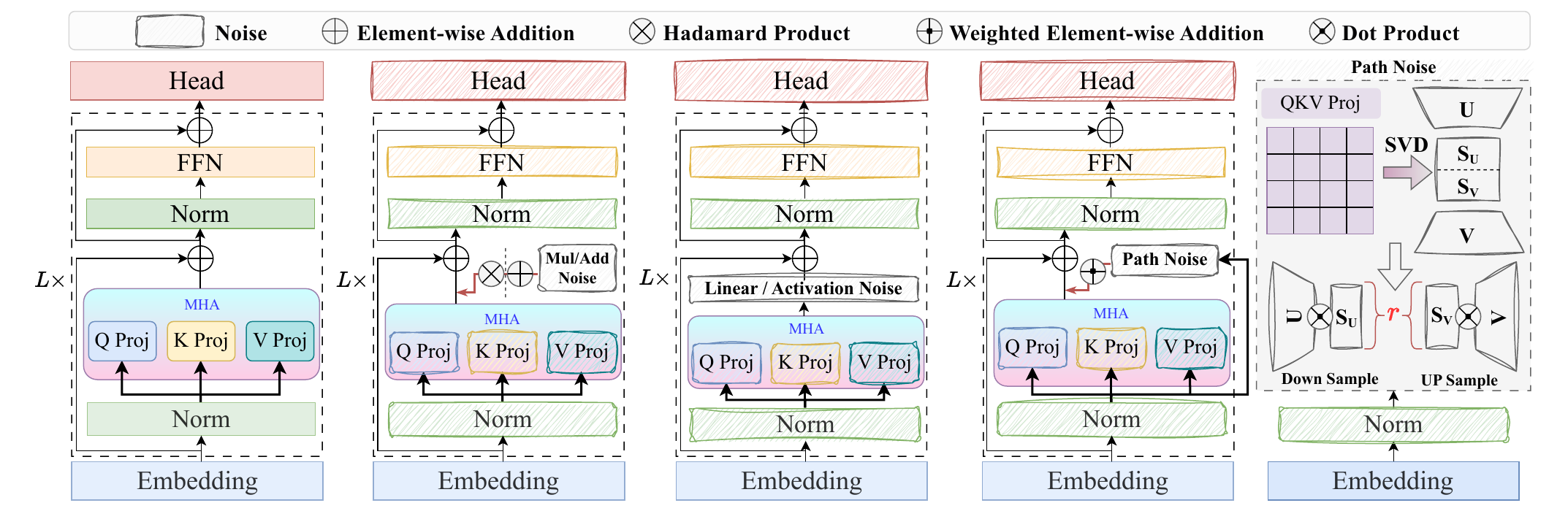}
\caption{Perturbation operators and injection points for pre-PEFT probing, including element-wise and activation perturbations.}
\label{fig:main:pert_overview}
\end{figure*}

\section{Datasets and Perturbation Settings}
We study pre-PEFT layer selection in the vision encoder. Our hypothesis is that robustness to controlled perturbations before fine-tuning correlates with subsequent PEFT effectiveness. Accordingly, we probe each layer without updating the base model, and later compare the probing signals with layer-wise PEFT results.

\subsection{Datasets Preparation}
We select 48 datasets that cover diverse VLM capabilities (image captioning, OCR, chart and table understanding, scientific image question answering, math question answering, etc.), including Screen2Words \cite{wang2021screen2words}, OCR-VQA \cite{mishra2019ocr}, Chart2Text \cite{kantharaj2022chart}, CLEVR-Math \cite{lindstrom2022clevr}, and ScienceQA \cite{lu2022learn}. Since most downstream fine-tuning tasks have small data sizes and there is some overlap among datasets, the training distribution may be dominated by a few datasets. To more clearly observe the correlation between ``perturbation scenarios--PEFT methods'' and to better match real-world settings, we sample 27,000 pairs from 1.8 million $\langle image, text \rangle$ pairs to construct a training mixture for subsequent perturbation probing and PEFT training.

To evaluate model performance after PEFT, we choose seven benchmarks for comprehensive evaluation: MMStar \cite{chen2024we} as the primary benchmark for overall comparison and consistency checking of basic capabilities, along with six additional benchmarks---ChartQA \cite{masry2022chartqa}, OCR-VQA \cite{mishra2019ocr}, HallusionBench \cite{guan2024hallusionbench}, ScienceQA \cite{lu2022learn}, RealWorldQA \cite{xai_realworldqa_2024}, and MMBench \cite{liu2023mmbench}---to evaluate diverse downstream tasks (English dialogue, form understanding, OCR, hallucination detection, science question answering, real-world question answering, etc.).

\subsection{Perturbation Settings}\label{sec:perturbation_settings}
Figure~\ref{fig:main:pert_overview} summarizes these perturbation operators and injection points.

To align perturbation design with layer-wise controllability, we focus on five PEFT methods that can be naturally compared at the layer level: LoRA \cite{hu2021lora}, VeRA \cite{kopiczko2023vera}, IA3 \cite{liu2022few}, LayerNorm-Tuning, and Partial Fine-Tuning.

We design five perturbation methods: Additive Noise (AddNoise), Multiplicative Noise (MulNoise), Linear Noise (LinearNoise), Activation Noise (ActNoise), and Path Noise (PathNoise). AddNoise and MulNoise sample from a normal distribution and apply element-wise noise scaled by a noise weight; LinearNoise and ActNoise introduce perturbations via a Kaiming-initialized linear layer \cite{he2019rethinking} and ReLU \cite{glorot2011deep}; PathNoise explicitly simulates the LoRA-style ``added bypass'' structure: we perform Singular Value Decomposition (SVD) on the original QKV generation linear layer to obtain $U$, $S$, and $V$, and then construct a new QKV generation matrix by down-sampling with $U \otimes S_u$ and up-sampling with $S_v \otimes V$ (where $S_u=S_v=\sqrt{S}$). The final output is a weighted sum of the original path and the new path.

In preliminary tests, we find that inserting a trainable linear layer (LinearNoise) into the vision encoder collapses the VLM (all benchmark scores drop to 0.0); thus LinearNoise is omitted from the reported tables.

\paragraph{Formalization}
Let $e$ denote the input embedding of a target layer and $qkv(\cdot)$ denote the original QKV projection. We instantiate perturbations by modifying the QKV output:
\begin{equation}
\begin{aligned}
\textsc{AddNoise:}\quad & qkv(e) + \alpha \,\epsilon, \ \ \epsilon \sim \mathcal{N}(0, I) \\
\textsc{MulNoise:}\quad & qkv(e) \odot (1 + \alpha \,\epsilon), \ \ \epsilon \sim \mathcal{N}(0, I) \\
\textsc{ActNoise:}\quad & \phi\!\left(qkv(e)\right), \ \ \phi(\cdot)=\mathrm{ReLU}(W(\cdot)) \\
\textsc{PathNoise:}\quad & (1-\alpha)\,qkv(e) + \alpha\,qkv'(e)
\end{aligned}
\end{equation}
where $\alpha$ is the perturbation weight and $qkv'(\cdot)$ is the SVD-constructed low-rank bypass path. We measure robustness by the change of downstream scores under the same evaluation protocol: layers with smaller drops are considered more robust.

\paragraph{Layer selection by probing}
Given an adaptation budget of $k$ layers, we rank candidate layers on a held-out split $\mathcal{D}$ by a probing score computed from ActNoise and PathNoise drops (smaller is better). Concretely, we define:
\begin{equation}
R_\ell=\frac{1}{2}\big(d_{\ell,\mathrm{act}} + d_{\ell,\mathrm{path}}\big),\quad
d_{\ell,\mathrm{path}}=\mathbb{E}_{\alpha\in\mathcal{A}_{\mathrm{path}}}\!\left[s_0 - s_{\ell,\alpha}\right],
\end{equation}
where $s_0$ is the unperturbed score and $s_{\ell,\alpha}$ is the score after perturbing layer $\ell$ with strength $\alpha$. We then select the top-$k$ layers with the smallest $R_\ell$ for subsequent multi-layer joint PEFT.

\paragraph{Hyperparameter Settings (Two Backbones)}\label{sec:hyperparams}
\begin{itemize}
    \item \textbf{Backbones}: InternVL2-1B (24L ViT + MLP); Qwen2.5-VL-3B (32L ViT) \cite{bai2025qwen25vl}
    \item \textbf{Training}: lr=$4\mathrm{e}{-5}$, wd=0.01, warmup=0.03, bf16
    \item \textbf{LoRA}: $r\in\{32,128,512\}$, $\alpha$=256, dropout=0.05
    \item \textbf{Perturbation}: Add/Mul/Act: $\alpha$=0.05; PathNoise: base $\alpha$=0.5 (rank=32), sweep $\alpha$=0.1–1.0
    \item \textbf{Inference}: temp=1.0, max\_len=32768, sliding-window (Transformers 4.37.2)
\end{itemize}

% 表格已迁移到 experiments（tables/intern_bench_layer_noise.tex）

\FloatBarrier
\section{Experiments and Analysis}

\subsection{Distribution of Model Parameters}\label{sec:cond_def}
We analyze the Q/K/V projection matrices in the 24-layer InternVL2-1B and 32-layer Qwen2.5-VL-3B vision encoders. For coarse cross-backbone comparisons, we segment the encoders into contiguous 8-layer blocks and refer to higher-index blocks/layers as deeper.

To quantify differences in the weight distributions of the Q/K/V projection layers, we compute six metrics for each layer and each projection matrix: variance, $L_1$ norm, $L_2$ norm, Frobenius norm, infinity norm, and condition number. For each layer $\ell$, we compute the statistics on the raw projection weights $W_q^{(\ell)}, W_k^{(\ell)}, W_v^{(\ell)}$. We define the condition number as $\kappa(W)=\sigma_{\max}(W)/\sigma_{\min}(W)$, where $\sigma_{\max}$ and $\sigma_{\min}$ are the largest and smallest singular values of $W$; $\kappa(\cdot)$ is computed on the raw weights without any attention-time scaling (e.g., $1/\sqrt{d_k}$) or additional normalization/reparameterization. For visualization and cross-layer comparison, we apply per-metric min--max scaling across layers within the same backbone to map each metric to $[0,1]$.

\begin{figure}[t]
\centering
\includegraphics[width=\columnwidth]{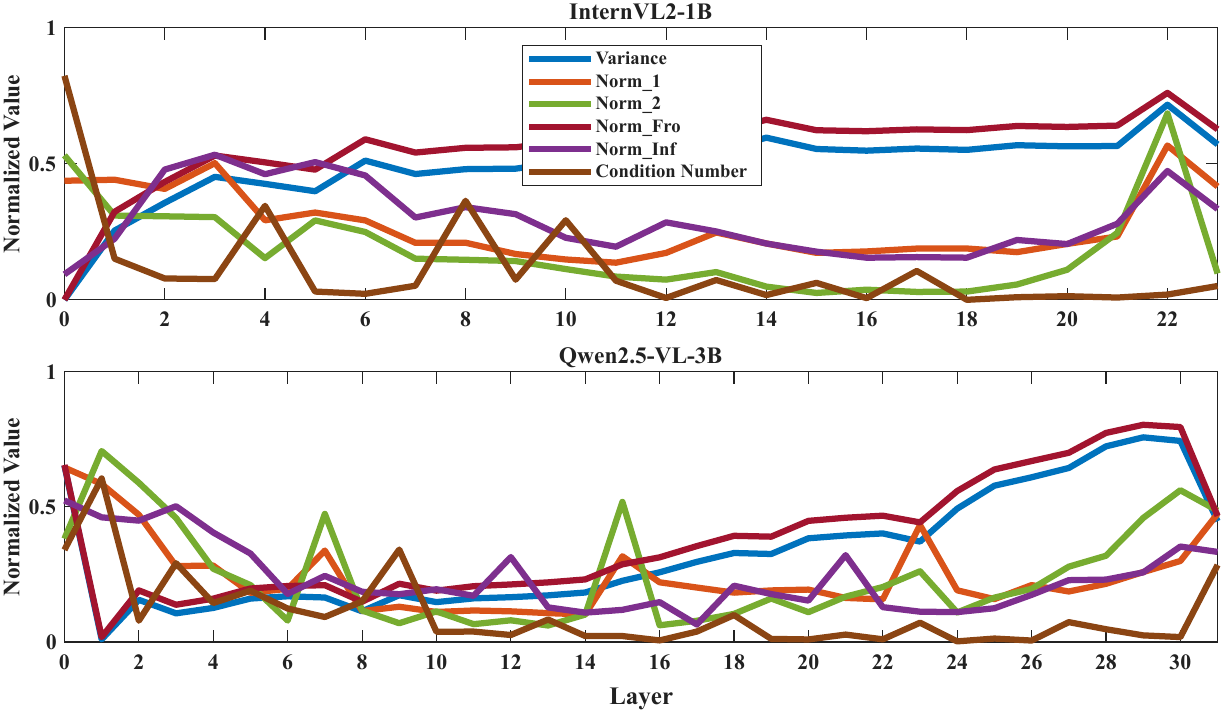}
\caption{Layer-wise statistics of the Q/K/V projection linear weights in the vision encoders of InternVL2-1B (top, 24 layers) and Qwen2.5-VL-3B (bottom, 32 layers). We report six metrics (variance, $L_1$, $L_2$, Frobenius, infinity norm, and condition number), each min--max normalized across layers within the same backbone.}
\label{fig:main:qkv_metrics}
\end{figure}

% 说明：IEEE 双栏下的 table* 往往只能浮到“后续页面页顶”。
% 将 Tables I--II 提前到 B 小节正文之前声明，可使其更自然地落在下一页页顶，
% 从而与随后引用它们的正文更接近，避免读者在 page 4/5 间来回翻找。
\begin{table*}[!t]
\caption{Block-level perturbation results on MMStar (left) and six benchmarks (right) for~\mbox{InternVL2-1B}. \ensuremath{\spadesuit}/\ensuremath{\clubsuit}/\ensuremath{\heartsuit}/\ensuremath{\diamondsuit}: column-best; \ensuremath{\uparrow}/\ensuremath{\downarrow}: vs.\ unperturbed.}
\label{tab:main:mmstar_part}
\centering
\setlength{\belowcaptionskip}{0pt}
\scriptsize
\begin{minipage}[b]{0.52\textwidth}\centering
\textbf{(a) MMStar (InternVL2-1B)}\\[-1pt]
\begin{adjustbox}{width=\linewidth}
\begin{tabular}[t]{@{}ccccccccc@{}}
\toprule
Layers & Perturbation & Overall & CP & FGP & IR & LR & MA & S\&T \\ \midrule
\multicolumn{1}{c|}{\multirow{4}{*}{\begin{tabular}[c]{@{}c@{}}0-7\\ (Block-1)\end{tabular}}} & \multicolumn{1}{c|}{PathNoise} & 37.61$_{\downarrow\text{8.06}}$ & 49.33$_{\downarrow\text{11.87}}$ & 28.22$_{\downarrow\text{7.38}}$ & 40.53$_{\downarrow\text{13.07}}$ & 27.96$_{\downarrow\text{7.24}}$ & $\spadesuit$44.13$_{\downarrow\text{3.87}}$ & 35.47$_{\downarrow\text{4.93}}$ \\
\multicolumn{1}{c|}{} & \multicolumn{1}{c|}{ActNoise} & 40.80$_{\downarrow\text{4.87}}$ & 54.40$_{\downarrow\text{6.80}}$ & 32.40$_{\downarrow\text{3.20}}$ & $\clubsuit$48.00$_{\downarrow\text{5.60}}$ & 31.20$_{\downarrow\text{4.00}}$ & 44.00$_{\downarrow\text{4.00}}$ & $\clubsuit$34.80$_{\downarrow\text{5.60}}$ \\
\multicolumn{1}{c|}{} & \multicolumn{1}{c|}{AddNoise} & 45.07$_{\downarrow\text{0.60}}$ & 61.20$_{\text{0.00}}$ & 33.60$_{\downarrow\text{2.00}}$ & 53.20$_{\downarrow\text{0.40}}$ & 34.80$_{\downarrow\text{0.40}}$ & 47.60$_{\downarrow\text{0.40}}$ & 40.00$_{\downarrow\text{0.40}}$ \\
\multicolumn{1}{c|}{} & \multicolumn{1}{c|}{MulNoise} & 45.13$_{\downarrow\text{0.53}}$ & 61.20$_{\text{0.00}}$ & 34.80$_{\downarrow\text{0.80}}$ & 52.80$_{\downarrow\text{0.80}}$ & 33.60$_{\downarrow\text{1.60}}$ & $\diamondsuit$48.80$_{\uparrow\text{0.80}}$ & 39.60$_{\downarrow\text{0.80}}$ \\
\midrule
\multicolumn{1}{c|}{\multirow{4}{*}{\begin{tabular}[c]{@{}c@{}}8-15\\ (Block-2)\end{tabular}}} & \multicolumn{1}{c|}{PathNoise} & 36.12$_{\downarrow\text{9.55}}$ & 44.22$_{\downarrow\text{16.98}}$ & 28.98$_{\downarrow\text{6.62}}$ & 39.07$_{\downarrow\text{14.53}}$ & 27.69$_{\downarrow\text{7.51}}$ & 43.78$_{\downarrow\text{4.22}}$ & 32.98$_{\downarrow\text{7.42}}$ \\
\multicolumn{1}{c|}{} & \multicolumn{1}{c|}{ActNoise} & 29.87$_{\downarrow\text{15.80}}$ & 37.20$_{\downarrow\text{24.00}}$ & 22.00$_{\downarrow\text{13.60}}$ & 29.60$_{\downarrow\text{24.00}}$ & 20.40$_{\downarrow\text{14.80}}$ & 40.40$_{\downarrow\text{7.60}}$ & 29.60$_{\downarrow\text{10.80}}$ \\
\multicolumn{1}{c|}{} & \multicolumn{1}{c|}{AddNoise} & 45.33$_{\downarrow\text{0.33}}$ & 61.20$_{\text{0.00}}$ & 34.00$_{\downarrow\text{1.60}}$ & 53.20$_{\downarrow\text{0.40}}$ & 35.20$_{\text{0.00}}$ & $\heartsuit$48.40$_{\uparrow\text{0.40}}$ & 40.00$_{\downarrow\text{0.40}}$ \\
\multicolumn{1}{c|}{} & \multicolumn{1}{c|}{MulNoise} & 45.40$_{\downarrow\text{0.27}}$ & 61.60$_{\uparrow\text{0.40}}$ & $\diamondsuit$35.20$_{\downarrow\text{0.40}}$ & 52.80$_{\downarrow\text{0.80}}$ & 34.80$_{\downarrow\text{0.40}}$ & 47.20$_{\downarrow\text{0.80}}$ & $\diamondsuit$40.80$_{\uparrow\text{0.40}}$ \\
\midrule
\multicolumn{1}{c|}{\multirow{4}{*}{\begin{tabular}[c]{@{}c@{}}16-23\\ (Block-3)\end{tabular}}} & \multicolumn{1}{c|}{PathNoise} & $\spadesuit$39.88$_{\downarrow\text{5.79}}$ & $\spadesuit$54.13$_{\downarrow\text{7.07}}$ & $\spadesuit$30.44$_{\downarrow\text{5.16}}$ & $\spadesuit$45.07$_{\downarrow\text{8.53}}$ & $\spadesuit$30.13$_{\downarrow\text{5.07}}$ & 43.20$_{\downarrow\text{4.80}}$ & $\spadesuit$36.31$_{\downarrow\text{4.09}}$ \\
\multicolumn{1}{c|}{} & \multicolumn{1}{c|}{ActNoise} & $\clubsuit$41.13$_{\downarrow\text{4.53}}$ & $\clubsuit$56.00$_{\downarrow\text{5.20}}$ & $\clubsuit$33.60$_{\downarrow\text{2.00}}$ & 45.20$_{\downarrow\text{8.40}}$ & $\clubsuit$34.00$_{\downarrow\text{1.20}}$ & $\clubsuit$44.80$_{\downarrow\text{3.20}}$ & 33.20$_{\downarrow\text{7.20}}$ \\
\multicolumn{1}{c|}{} & \multicolumn{1}{c|}{AddNoise} & $\heartsuit$45.60$_{\downarrow\text{0.07}}$ & $\heartsuit$61.60$_{\uparrow\text{0.40}}$ & $\heartsuit$34.80$_{\downarrow\text{0.80}}$ & $\heartsuit$53.60$_{\text{0.00}}$ & $\heartsuit$36.00$_{\uparrow\text{0.80}}$ & 47.20$_{\downarrow\text{0.80}}$ & $\heartsuit$40.40$_{\text{0.00}}$ \\
\multicolumn{1}{c|}{} & \multicolumn{1}{c|}{MulNoise} & $\diamondsuit$45.53$_{\downarrow\text{0.13}}$ & $\diamondsuit$62.00$_{\uparrow\text{0.80}}$ & 34.80$_{\downarrow\text{0.80}}$ & $\diamondsuit$53.60$_{\text{0.00}}$ & $\diamondsuit$35.60$_{\uparrow\text{0.40}}$ & 47.60$_{\downarrow\text{0.40}}$ & 39.60$_{\downarrow\text{0.80}}$ \\
\bottomrule
\end{tabular}
\end{adjustbox}
\end{minipage}\hfill
\begin{minipage}[b]{0.46\textwidth}\centering
\textbf{(b) Benchmarks (InternVL2-1B)}\\[-1pt]
\begin{adjustbox}{width=\linewidth}
\begin{tabular}[t]{@{}cccccccc@{}}
\toprule
Layers & Perturbation & CQA & HB & MMB & OCRVQA & RWQA & SQA \\ \midrule
\multicolumn{1}{c|}{\multirow{4}{*}{\begin{tabular}[c]{@{}c@{}}0-7\\ (Block-1)\end{tabular}}} & \multicolumn{1}{c|}{PathNoise} & 28.79$_{\downarrow\text{29.09}}$ & 29.59$_{\downarrow\text{5.62}}$ & 47.69$_{\downarrow\text{19.49}}$ & 31.98$_{\downarrow\text{11.53}}$ & $\spadesuit$42.99$_{\downarrow\text{5.77}}$ & 78.83$_{\downarrow\text{8.58}}$ \\
\multicolumn{1}{c|}{} & \multicolumn{1}{c|}{ActNoise} & 27.84$_{\downarrow\text{30.04}}$ & $\clubsuit$30.82$_{\downarrow\text{4.39}}$ & $\clubsuit$60.91$_{\downarrow\text{6.27}}$ & $\clubsuit$35.37$_{\downarrow\text{8.13}}$ & $\clubsuit$44.31$_{\downarrow\text{4.44}}$ & 82.30$_{\downarrow\text{5.11}}$ \\
\multicolumn{1}{c|}{} & \multicolumn{1}{c|}{AddNoise} & $\heartsuit$58.16$_{\uparrow\text{0.28}}$ & 35.49$_{\uparrow\text{0.29}}$ & $\heartsuit$67.35$_{\uparrow\text{0.17}}$ & $\heartsuit$43.54$_{\uparrow\text{0.03}}$ & $\heartsuit$48.76$_{\text{0.00}}$ & $\heartsuit$87.36$_{\downarrow\text{0.05}}$ \\
\multicolumn{1}{c|}{} & \multicolumn{1}{c|}{MulNoise} & 58.04$_{\uparrow\text{0.16}}$ & 31.27$_{\downarrow\text{3.93}}$ & 67.18$_{\text{0.00}}$ & 43.51$_{\text{0.00}}$ & $\diamondsuit$48.89$_{\uparrow\text{0.13}}$ & 87.26$_{\downarrow\text{0.15}}$ \\
\midrule
\multicolumn{1}{c|}{\multirow{4}{*}{\begin{tabular}[c]{@{}c@{}}8-15\\ (Block-2)\end{tabular}}} & \multicolumn{1}{c|}{PathNoise} & 24.57$_{\downarrow\text{33.31}}$ & $\spadesuit$30.52$_{\downarrow\text{4.68}}$ & 42.50$_{\downarrow\text{24.68}}$ & 27.96$_{\downarrow\text{15.55}}$ & 42.67$_{\downarrow\text{6.09}}$ & 78.09$_{\downarrow\text{9.32}}$ \\
\multicolumn{1}{c|}{} & \multicolumn{1}{c|}{ActNoise} & 12.88$_{\downarrow\text{45.00}}$ & 26.22$_{\downarrow\text{8.98}}$ & 28.69$_{\downarrow\text{38.49}}$ & 20.64$_{\downarrow\text{22.87}}$ & 37.52$_{\downarrow\text{11.24}}$ & 73.18$_{\downarrow\text{14.23}}$ \\
\multicolumn{1}{c|}{} & \multicolumn{1}{c|}{AddNoise} & 57.96$_{\uparrow\text{0.08}}$ & $\heartsuit$36.14$_{\uparrow\text{0.94}}$ & 67.18$_{\text{0.00}}$ & 43.53$_{\uparrow\text{0.03}}$ & 48.24$_{\downarrow\text{0.52}}$ & 87.31$_{\downarrow\text{0.10}}$ \\
\multicolumn{1}{c|}{} & \multicolumn{1}{c|}{MulNoise} & $\diamondsuit$58.08$_{\uparrow\text{0.20}}$ & 35.10$_{\downarrow\text{0.10}}$ & $\diamondsuit$67.70$_{\uparrow\text{0.52}}$ & $\diamondsuit$43.54$_{\uparrow\text{0.04}}$ & 47.84$_{\downarrow\text{0.92}}$ & $\diamondsuit$87.46$_{\uparrow\text{0.05}}$ \\
\midrule
\multicolumn{1}{c|}{\multirow{4}{*}{\begin{tabular}[c]{@{}c@{}}16-23\\ (Block-3)\end{tabular}}} & \multicolumn{1}{c|}{PathNoise} & $\spadesuit$36.39$_{\downarrow\text{21.49}}$ & 29.77$_{\downarrow\text{5.44}}$ & $\spadesuit$53.00$_{\downarrow\text{14.18}}$ & $\spadesuit$33.74$_{\downarrow\text{9.77}}$ & 42.95$_{\downarrow\text{5.81}}$ & $\spadesuit$82.23$_{\downarrow\text{5.17}}$ \\
\multicolumn{1}{c|}{} & \multicolumn{1}{c|}{ActNoise} & $\clubsuit$49.52$_{\downarrow\text{8.36}}$ & 30.41$_{\downarrow\text{4.79}}$ & 51.46$_{\downarrow\text{15.72}}$ & 34.76$_{\downarrow\text{8.74}}$ & $\clubsuit$44.31$_{\downarrow\text{4.44}}$ & $\clubsuit$82.90$_{\downarrow\text{4.51}}$ \\
\multicolumn{1}{c|}{} & \multicolumn{1}{c|}{AddNoise} & $\heartsuit$58.16$_{\uparrow\text{0.28}}$ & 31.56$_{\downarrow\text{3.64}}$ & 67.27$_{\uparrow\text{0.09}}$ & 43.50$_{\downarrow\text{0.01}}$ & 48.63$_{\downarrow\text{0.13}}$ & $\heartsuit$87.36$_{\downarrow\text{0.05}}$ \\
\multicolumn{1}{c|}{} & \multicolumn{1}{c|}{MulNoise} & 57.76$_{\downarrow\text{0.12}}$ & $\diamondsuit$35.68$_{\uparrow\text{0.47}}$ & 67.27$_{\uparrow\text{0.09}}$ & 43.50$_{\text{0.00}}$ & 48.50$_{\downarrow\text{0.26}}$ & 87.41$_{\text{0.00}}$ \\
\bottomrule
\end{tabular}
\end{adjustbox}
\end{minipage}
\end{table*}

\begin{table*}[!t]
\caption{Block-level perturbation results on MMStar (left) and six benchmarks (right) for~\mbox{Qwen2.5-VL-3B}. \ensuremath{\spadesuit}/\ensuremath{\clubsuit}/\ensuremath{\heartsuit}/\ensuremath{\diamondsuit}: column-best; \ensuremath{\uparrow}/\ensuremath{\downarrow}: vs.\ unperturbed.}
\label{tab:main:qwen_mmstar_part}
\centering
\setlength{\belowcaptionskip}{0pt}
\scriptsize
\begin{minipage}[b]{0.52\textwidth}\centering
\textbf{(a) MMStar (Qwen2.5-VL-3B)}\\[-1pt]
\begin{adjustbox}{width=\linewidth}
\begin{tabular}[t]{@{}ccccccccc@{}}
\toprule
Layers & Perturbation & Overall & CP & FGP & IR & LR & MA & S\&T \\ \midrule
\multicolumn{1}{c|}{\multirow{4}{*}{\begin{tabular}[c]{@{}c@{}}0-7\\ (Block-1)\end{tabular}}} & \multicolumn{1}{c|}{PathNoise} & $\spadesuit$54.64$_{\downarrow\text{0.26}}$ & $\spadesuit$66.40$_{\downarrow\text{1.20}}$ & 46.40$_{\uparrow\text{0.40}}$ & 60.80$_{\uparrow\text{0.40}}$ & $\spadesuit$57.43$_{\uparrow\text{2.41}}$ & $\spadesuit$60.80$_{\downarrow\text{1.60}}$ & 36.00$_{\downarrow\text{2.00}}$ \\
\multicolumn{1}{c|}{} & \multicolumn{1}{c|}{ActNoise} & $\clubsuit$55.57$_{\uparrow\text{0.67}}$ & $\clubsuit$68.00$_{\uparrow\text{0.40}}$ & 46.00$_{\text{0.00}}$ & 60.40$_{\text{0.00}}$ & $\clubsuit$57.43$_{\uparrow\text{2.41}}$ & $\clubsuit$63.20$_{\uparrow\text{0.80}}$ & 38.40$_{\uparrow\text{0.40}}$ \\
\multicolumn{1}{c|}{} & \multicolumn{1}{c|}{AddNoise} & 55.30$_{\uparrow\text{0.40}}$ & 67.60$_{\text{0.00}}$ & 44.80$_{\downarrow\text{1.20}}$ & 60.00$_{\downarrow\text{0.40}}$ & $\heartsuit$57.43$_{\uparrow\text{2.41}}$ & $\heartsuit$63.60$_{\uparrow\text{1.20}}$ & 38.40$_{\uparrow\text{0.40}}$ \\
\multicolumn{1}{c|}{} & \multicolumn{1}{c|}{MulNoise} & 55.10$_{\uparrow\text{0.20}}$ & $\diamondsuit$67.60$_{\text{0.00}}$ & $\diamondsuit$47.60$_{\uparrow\text{1.60}}$ & 60.40$_{\text{0.00}}$ & 55.42$_{\uparrow\text{0.40}}$ & 61.60$_{\downarrow\text{0.80}}$ & 38.00$_{\text{0.00}}$ \\
\midrule
\multicolumn{1}{c|}{\multirow{4}{*}{\begin{tabular}[c]{@{}c@{}}8-15\\ (Block-2)\end{tabular}}} & \multicolumn{1}{c|}{PathNoise} & 53.57$_{\downarrow\text{1.33}}$ & 65.20$_{\downarrow\text{2.40}}$ & 46.00$_{\text{0.00}}$ & 58.80$_{\downarrow\text{1.60}}$ & 54.62$_{\downarrow\text{0.40}}$ & 60.80$_{\downarrow\text{1.60}}$ & 36.00$_{\downarrow\text{2.00}}$ \\
\multicolumn{1}{c|}{} & \multicolumn{1}{c|}{ActNoise} & 55.44$_{\uparrow\text{0.54}}$ & 68.00$_{\uparrow\text{0.40}}$ & $\clubsuit$46.40$_{\uparrow\text{0.40}}$ & 60.80$_{\uparrow\text{0.40}}$ & 57.03$_{\uparrow\text{2.01}}$ & 62.00$_{\downarrow\text{0.40}}$ & 38.40$_{\uparrow\text{0.40}}$ \\
\multicolumn{1}{c|}{} & \multicolumn{1}{c|}{AddNoise} & $\heartsuit$55.97$_{\uparrow\text{1.07}}$ & $\heartsuit$68.00$_{\uparrow\text{0.40}}$ & $\heartsuit$47.60$_{\uparrow\text{1.60}}$ & 61.20$_{\uparrow\text{0.80}}$ & 56.22$_{\uparrow\text{1.20}}$ & 62.80$_{\uparrow\text{0.40}}$ & $\heartsuit$40.00$_{\uparrow\text{2.00}}$ \\
\multicolumn{1}{c|}{} & \multicolumn{1}{c|}{MulNoise} & 55.24$_{\uparrow\text{0.34}}$ & 65.60$_{\downarrow\text{2.00}}$ & 46.00$_{\text{0.00}}$ & 60.80$_{\uparrow\text{0.40}}$ & $\diamondsuit$57.83$_{\uparrow\text{2.81}}$ & $\diamondsuit$62.00$_{\downarrow\text{0.40}}$ & 39.20$_{\uparrow\text{1.20}}$ \\
\midrule
\multicolumn{1}{c|}{\multirow{4}{*}{\begin{tabular}[c]{@{}c@{}}16-23\\ (Block-3)\end{tabular}}} & \multicolumn{1}{c|}{PathNoise} & 52.30$_{\downarrow\text{2.60}}$ & 65.20$_{\downarrow\text{2.40}}$ & 44.80$_{\downarrow\text{1.20}}$ & 60.40$_{\text{0.00}}$ & 52.61$_{\downarrow\text{2.41}}$ & 58.40$_{\downarrow\text{4.00}}$ & 32.40$_{\downarrow\text{5.60}}$ \\
\multicolumn{1}{c|}{} & \multicolumn{1}{c|}{ActNoise} & 55.10$_{\uparrow\text{0.20}}$ & 67.20$_{\downarrow\text{0.40}}$ & 45.20$_{\downarrow\text{0.80}}$ & $\clubsuit$61.60$_{\uparrow\text{1.20}}$ & 55.02$_{\text{0.00}}$ & 62.80$_{\uparrow\text{0.40}}$ & 38.80$_{\uparrow\text{0.80}}$ \\
\multicolumn{1}{c|}{} & \multicolumn{1}{c|}{AddNoise} & 55.37$_{\uparrow\text{0.47}}$ & 68.00$_{\uparrow\text{0.40}}$ & 45.20$_{\downarrow\text{0.80}}$ & $\heartsuit$61.60$_{\uparrow\text{1.20}}$ & 55.82$_{\uparrow\text{0.80}}$ & 61.60$_{\downarrow\text{0.80}}$ & 40.00$_{\uparrow\text{2.00}}$ \\
\multicolumn{1}{c|}{} & \multicolumn{1}{c|}{MulNoise} & 54.57$_{\downarrow\text{0.33}}$ & 66.80$_{\downarrow\text{0.80}}$ & 46.40$_{\uparrow\text{0.40}}$ & 60.00$_{\downarrow\text{0.40}}$ & 55.02$_{\text{0.00}}$ & 61.20$_{\downarrow\text{1.20}}$ & 38.00$_{\text{0.00}}$ \\
\midrule
\multicolumn{1}{c|}{\multirow{4}{*}{\begin{tabular}[c]{@{}c@{}}24-31\\ (Block-4)\end{tabular}}} & \multicolumn{1}{c|}{PathNoise} & 53.77$_{\downarrow\text{1.13}}$ & 64.80$_{\downarrow\text{2.80}}$ & $\spadesuit$46.80$_{\uparrow\text{0.80}}$ & $\spadesuit$62.00$_{\uparrow\text{1.60}}$ & 52.61$_{\downarrow\text{2.41}}$ & 59.60$_{\downarrow\text{2.80}}$ & $\spadesuit$36.80$_{\downarrow\text{1.20}}$ \\
\multicolumn{1}{c|}{} & \multicolumn{1}{c|}{ActNoise} & 55.17$_{\uparrow\text{0.27}}$ & 67.60$_{\text{0.00}}$ & 46.00$_{\text{0.00}}$ & 61.20$_{\uparrow\text{0.80}}$ & 54.62$_{\downarrow\text{0.40}}$ & 61.60$_{\downarrow\text{0.80}}$ & $\clubsuit$40.00$_{\uparrow\text{2.00}}$ \\
\multicolumn{1}{c|}{} & \multicolumn{1}{c|}{AddNoise} & 55.17$_{\uparrow\text{0.27}}$ & 67.20$_{\downarrow\text{0.40}}$ & 45.20$_{\downarrow\text{0.80}}$ & 61.60$_{\uparrow\text{1.20}}$ & 54.62$_{\downarrow\text{0.40}}$ & 62.80$_{\uparrow\text{0.40}}$ & 39.60$_{\uparrow\text{1.60}}$ \\
\multicolumn{1}{c|}{} & \multicolumn{1}{c|}{MulNoise} & $\diamondsuit$55.37$_{\uparrow\text{0.47}}$ & 66.80$_{\downarrow\text{0.80}}$ & 46.80$_{\uparrow\text{0.80}}$ & $\diamondsuit$61.20$_{\uparrow\text{0.80}}$ & 55.82$_{\uparrow\text{0.80}}$ & 62.00$_{\downarrow\text{0.40}}$ & $\diamondsuit$39.60$_{\uparrow\text{1.60}}$ \\
\bottomrule
\end{tabular}

\end{adjustbox}
\end{minipage}\hfill
\begin{minipage}[b]{0.46\textwidth}\centering
\textbf{(b) Benchmarks (Qwen2.5-VL-3B)}\\[-1pt]
\begin{adjustbox}{width=\linewidth}
\begin{tabular}[t]{@{}cccccccc@{}}
\toprule
Layers & Perturbation & CQA & HB & MMB & OCRVQA & RWQA & SQA \\ \midrule
\multicolumn{1}{c|}{\multirow{4}{*}{\begin{tabular}[c]{@{}c@{}}0-7\\ (Block-1)\end{tabular}}} & \multicolumn{1}{c|}{PathNoise} & $\spadesuit$84.28$_{\uparrow\text{0.20}}$ & $\spadesuit$61.09$_{\uparrow\text{0.73}}$ & $\spadesuit$82.27$_{\downarrow\text{0.16}}$ & $\spadesuit$76.80$_{\downarrow\text{0.50}}$ & $\spadesuit$65.23$_{\downarrow\text{1.18}}$ & 74.81$_{\downarrow\text{0.75}}$ \\
\multicolumn{1}{c|}{} & \multicolumn{1}{c|}{ActNoise} & 84.12$_{\uparrow\text{0.04}}$ & 60.15$_{\downarrow\text{0.21}}$ & $\clubsuit$82.53$_{\uparrow\text{0.10}}$ & 77.30$_{\text{0.00}}$ & $\clubsuit$65.75$_{\downarrow\text{0.66}}$ & $\clubsuit$75.46$_{\downarrow\text{0.10}}$ \\
\multicolumn{1}{c|}{} & \multicolumn{1}{c|}{AddNoise} & $\heartsuit$84.16$_{\uparrow\text{0.08}}$ & $\heartsuit$61.20$_{\uparrow\text{0.84}}$ & $\heartsuit$82.59$_{\uparrow\text{0.16}}$ & 77.00$_{\downarrow\text{0.30}}$ & 65.49$_{\downarrow\text{0.92}}$ & 75.06$_{\downarrow\text{0.50}}$ \\
\multicolumn{1}{c|}{} & \multicolumn{1}{c|}{MulNoise} & 83.80$_{\downarrow\text{0.28}}$ & $\diamondsuit$60.78$_{\uparrow\text{0.42}}$ & 82.53$_{\uparrow\text{0.10}}$ & 76.90$_{\downarrow\text{0.40}}$ & 65.23$_{\downarrow\text{1.18}}$ & 75.16$_{\downarrow\text{0.40}}$ \\
\midrule
\multicolumn{1}{c|}{\multirow{4}{*}{\begin{tabular}[c]{@{}c@{}}8-15\\ (Block-2)\end{tabular}}} & \multicolumn{1}{c|}{PathNoise} & 83.48$_{\downarrow\text{0.60}}$ & 60.46$_{\uparrow\text{0.10}}$ & 80.89$_{\downarrow\text{1.54}}$ & 76.60$_{\downarrow\text{0.70}}$ & 63.53$_{\downarrow\text{2.88}}$ & 74.71$_{\downarrow\text{0.85}}$ \\
\multicolumn{1}{c|}{} & \multicolumn{1}{c|}{ActNoise} & 84.08$_{\text{0.00}}$ & 59.73$_{\downarrow\text{0.63}}$ & 82.20$_{\downarrow\text{0.23}}$ & 77.20$_{\downarrow\text{0.10}}$ & 65.23$_{\downarrow\text{1.18}}$ & 75.01$_{\downarrow\text{0.55}}$ \\
\multicolumn{1}{c|}{} & \multicolumn{1}{c|}{AddNoise} & 84.04$_{\downarrow\text{0.04}}$ & 60.78$_{\uparrow\text{0.42}}$ & 82.41$_{\downarrow\text{0.02}}$ & 76.70$_{\downarrow\text{0.60}}$ & $\heartsuit$65.88$_{\downarrow\text{0.53}}$ & 75.11$_{\downarrow\text{0.45}}$ \\
\multicolumn{1}{c|}{} & \multicolumn{1}{c|}{MulNoise} & 83.84$_{\downarrow\text{0.24}}$ & 60.67$_{\uparrow\text{0.31}}$ & 82.74$_{\uparrow\text{0.31}}$ & 77.10$_{\downarrow\text{0.20}}$ & 65.10$_{\downarrow\text{1.31}}$ & 74.91$_{\downarrow\text{0.65}}$ \\
\midrule
\multicolumn{1}{c|}{\multirow{4}{*}{\begin{tabular}[c]{@{}c@{}}16-23\\ (Block-3)\end{tabular}}} & \multicolumn{1}{c|}{PathNoise} & 82.44$_{\downarrow\text{1.64}}$ & 60.15$_{\downarrow\text{0.21}}$ & 81.63$_{\downarrow\text{0.80}}$ & 75.40$_{\downarrow\text{1.90}}$ & 63.40$_{\downarrow\text{3.01}}$ & 74.47$_{\downarrow\text{1.09}}$ \\
\multicolumn{1}{c|}{} & \multicolumn{1}{c|}{ActNoise} & 84.12$_{\uparrow\text{0.04}}$ & $\clubsuit$60.57$_{\uparrow\text{0.21}}$ & 82.47$_{\uparrow\text{0.04}}$ & 76.60$_{\downarrow\text{0.70}}$ & 65.62$_{\downarrow\text{0.79}}$ & 75.26$_{\downarrow\text{0.30}}$ \\
\multicolumn{1}{c|}{} & \multicolumn{1}{c|}{AddNoise} & 83.96$_{\downarrow\text{0.12}}$ & 60.78$_{\uparrow\text{0.42}}$ & 82.33$_{\downarrow\text{0.10}}$ & 76.90$_{\downarrow\text{0.40}}$ & 65.62$_{\downarrow\text{0.79}}$ & $\heartsuit$75.21$_{\downarrow\text{0.35}}$ \\
\multicolumn{1}{c|}{} & \multicolumn{1}{c|}{MulNoise} & $\diamondsuit$84.00$_{\downarrow\text{0.08}}$ & 60.15$_{\downarrow\text{0.21}}$ & $\diamondsuit$82.78$_{\uparrow\text{0.35}}$ & 77.10$_{\downarrow\text{0.20}}$ & $\diamondsuit$66.41$_{\text{0.00}}$ & 75.16$_{\downarrow\text{0.40}}$ \\
\midrule
\multicolumn{1}{c|}{\multirow{4}{*}{\begin{tabular}[c]{@{}c@{}}24-31\\ (Block-4)\end{tabular}}} & \multicolumn{1}{c|}{PathNoise} & 82.16$_{\downarrow\text{1.92}}$ & 59.10$_{\downarrow\text{1.26}}$ & 81.96$_{\downarrow\text{0.47}}$ & 76.80$_{\downarrow\text{0.50}}$ & 62.48$_{\downarrow\text{3.93}}$ & $\spadesuit$74.91$_{\downarrow\text{0.65}}$ \\
\multicolumn{1}{c|}{} & \multicolumn{1}{c|}{ActNoise} & $\clubsuit$84.16$_{\uparrow\text{0.08}}$ & 60.57$_{\uparrow\text{0.21}}$ & 82.49$_{\uparrow\text{0.06}}$ & $\clubsuit$77.70$_{\uparrow\text{0.40}}$ & 65.62$_{\downarrow\text{0.79}}$ & 75.01$_{\downarrow\text{0.55}}$ \\
\multicolumn{1}{c|}{} & \multicolumn{1}{c|}{AddNoise} & 83.96$_{\downarrow\text{0.12}}$ & 60.78$_{\uparrow\text{0.42}}$ & 82.35$_{\downarrow\text{0.08}}$ & $\heartsuit$77.10$_{\downarrow\text{0.20}}$ & 65.88$_{\downarrow\text{0.53}}$ & 75.01$_{\downarrow\text{0.55}}$ \\
\multicolumn{1}{c|}{} & \multicolumn{1}{c|}{MulNoise} & 84.00$_{\downarrow\text{0.08}}$ & 60.46$_{\uparrow\text{0.10}}$ & 82.61$_{\uparrow\text{0.18}}$ & $\diamondsuit$77.20$_{\downarrow\text{0.10}}$ & 65.36$_{\downarrow\text{1.05}}$ & $\diamondsuit$75.31$_{\downarrow\text{0.25}}$ \\
\bottomrule
\end{tabular}

\end{adjustbox}
\end{minipage}
\end{table*}

In Figure \ref{fig:main:qkv_metrics}, deeper layers generally exhibit larger variance and norm-related metrics, while condition-number fluctuations are more concentrated in earlier layers.
% Middle layers often act as a transition, abstracting low-level perceptual information and preparing it for deeper layers to extract specific details. This analysis confirms that deeper layers are characterized by high variance in their parameter distributions.

\subsection{Segmented Analysis of Vision Encoder Layers}

To estimate promising PEFT layers before tuning, we inject perturbations into the predefined 8-layer blocks and evaluate MMStar plus six benchmarks; PathNoise is used as the LoRA-like perturbation. Tables \ref{tab:main:mmstar_part}(a) and \ref{tab:main:qwen_mmstar_part}(a) report the block-level MMStar results for InternVL2-1B and Qwen2.5-VL-3B, respectively.

Across both backbones, ActNoise and PathNoise produce the largest degradation, whereas AddNoise and MulNoise are much milder; moreover, block-level robustness is not monotonic with depth (Tables \ref{tab:main:mmstar_part} and \ref{tab:main:qwen_mmstar_part}). This motivates the finer layer-wise analysis below.

\subsection{Layer-Wise Analysis of Vision Encoder}
Table \ref{tab:main:mmstar_layer_noise} shows substantial within-block variation on InternVL2-1B, with several middle layers consistently more fragile under both ActNoise and PathNoise, suggesting layer-level PEFT selection rather than contiguous block selection.

\subsection{PEFT Analysis of Vision Encoder}
We evaluate layer-wise PEFT on InternVL2-1B with LoRA ($r=32/128/512$), VeRA, IA3, LayerNorm-Tuning, and Partial-Tuning, and on Qwen2.5-VL-3B with LoRA ($r=32$) plus the same baselines.

\begin{figure*}[!t]
\centering
% 说明：将 Fig. 5 与 Table IV 改回常规双栏浮动体，并放在 D 小节正文之后再声明，
% 让它们顺延到后续页顶，避免 strip 在当前版面下发生跨页截断。
% 按“固定子图宽度盒子 + 固定高度”排版，避免各 PDF 画布不一致导致视觉大小/对齐不齐。
\newcommand{\ijcnnradarcell}[3]{%
    \begin{minipage}[t]{0.24\textwidth}
    \centering
    \includegraphics[height=3.15cm,keepaspectratio]{#1}\\[-0.35ex]
    {\footnotesize (#2) #3}
    \end{minipage}%
}
\ijcnnradarcell{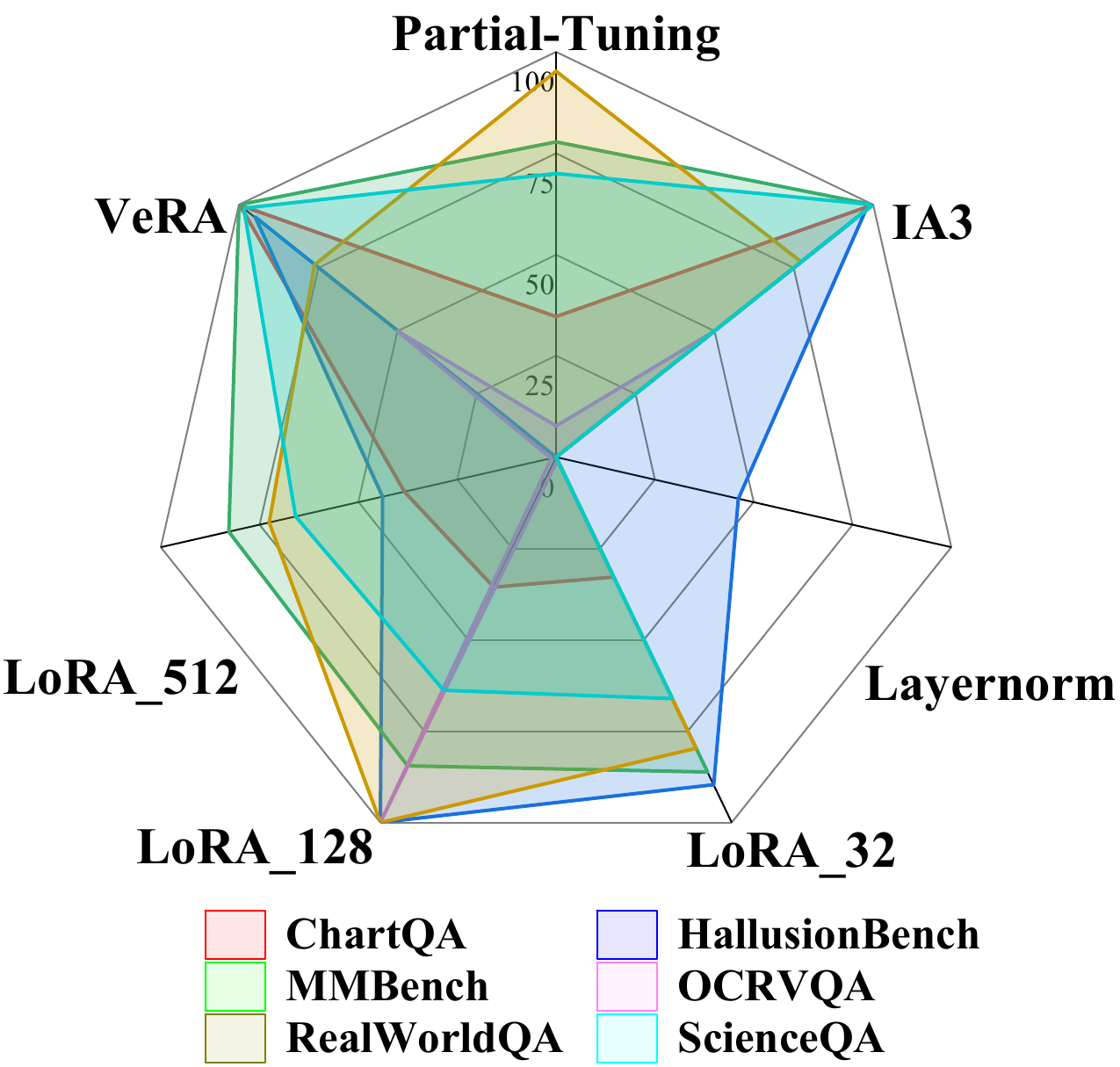}{a}{All}\hfill
\ijcnnradarcell{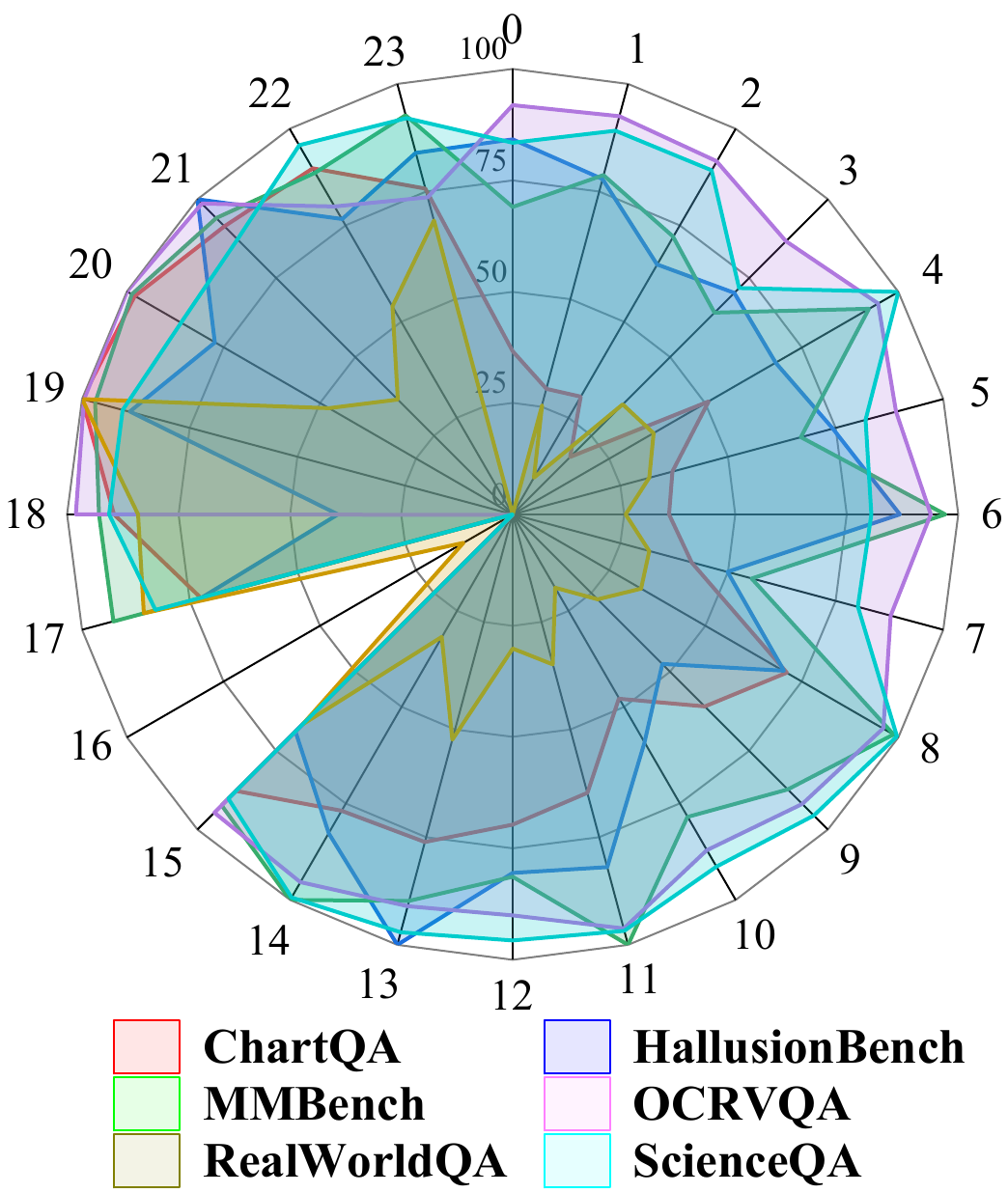}{b}{Partial}\hfill
\ijcnnradarcell{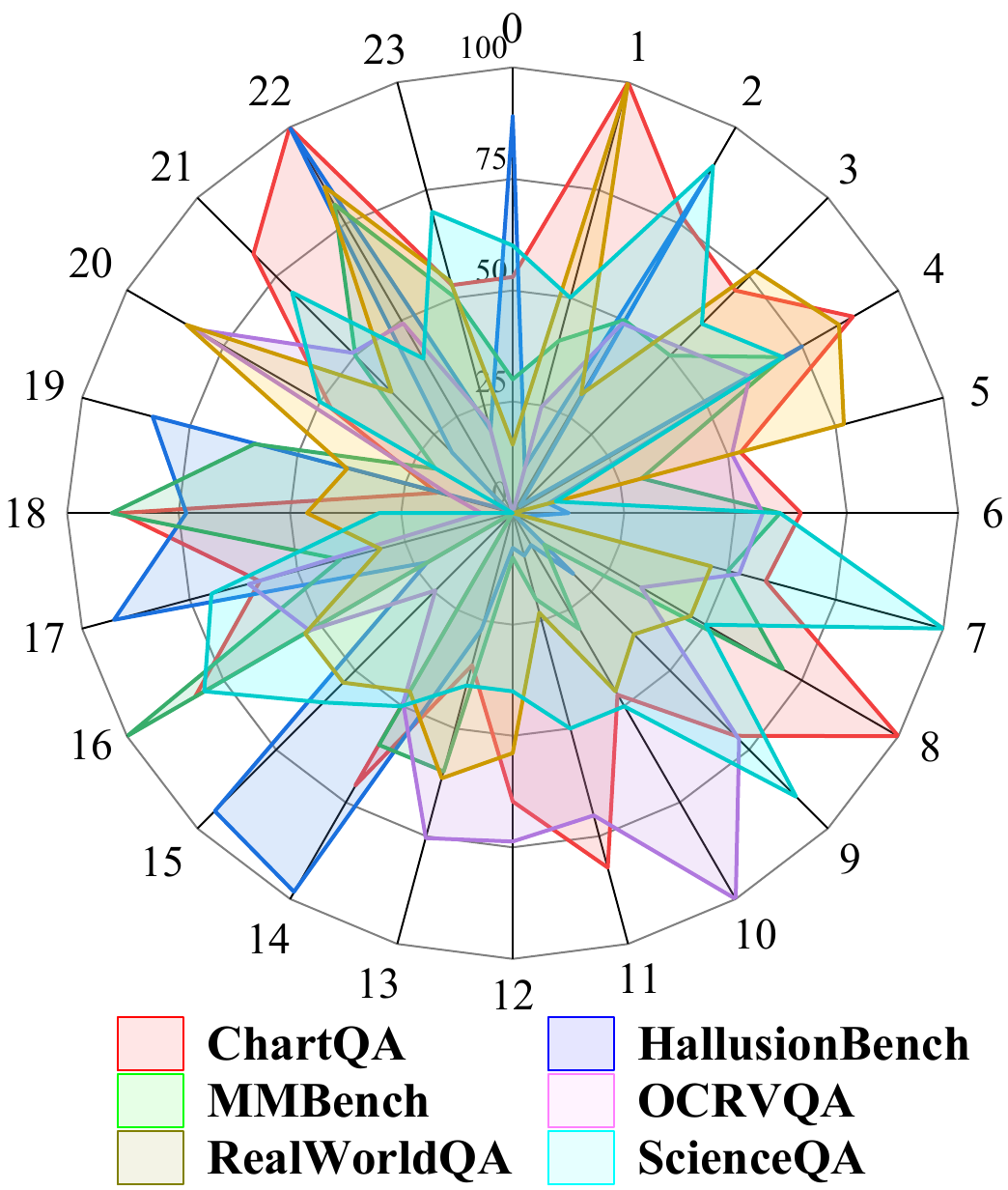}{c}{IA3}\hfill
\ijcnnradarcell{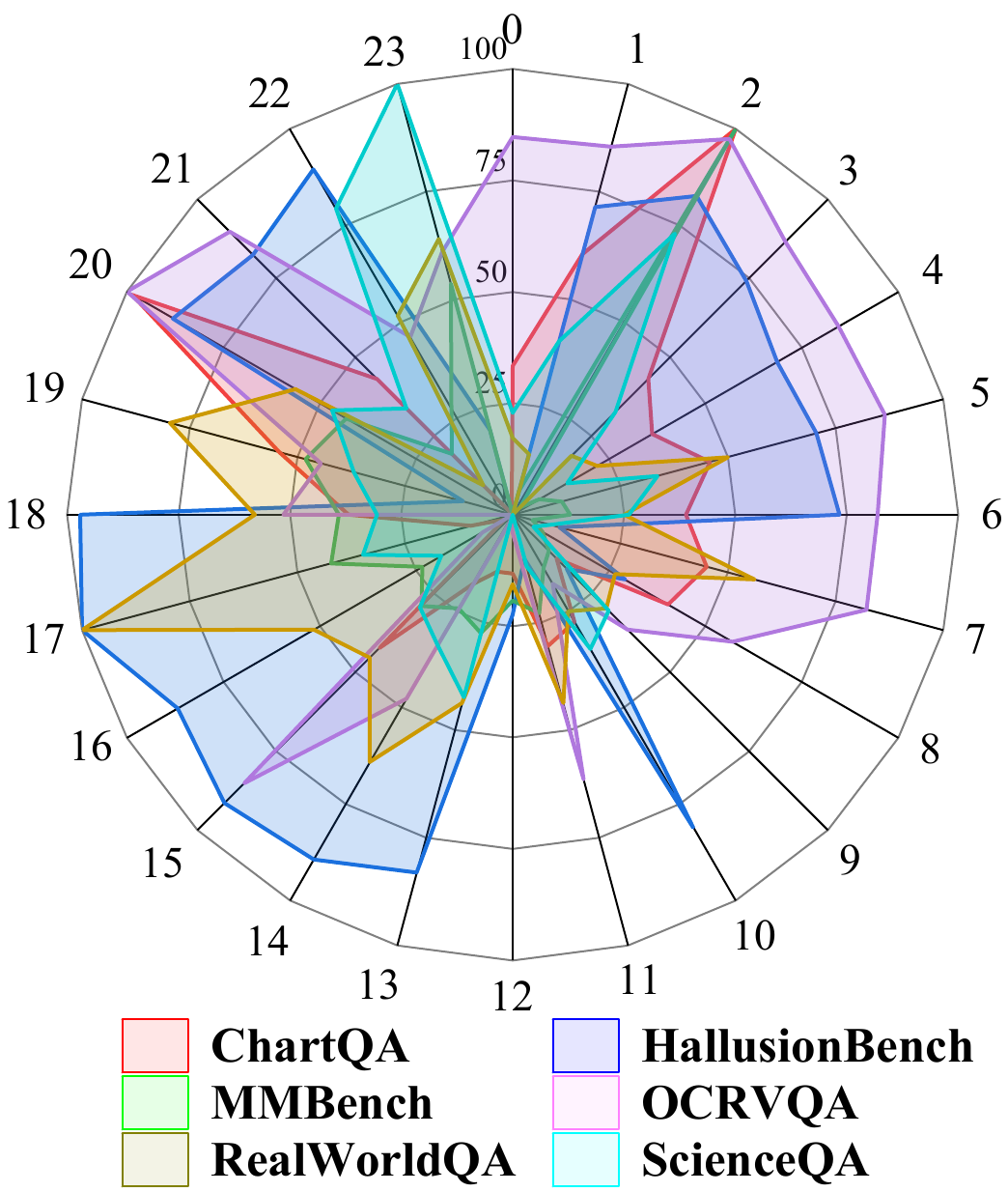}{d}{LayerNorm}\\[-0.6ex]
\ijcnnradarcell{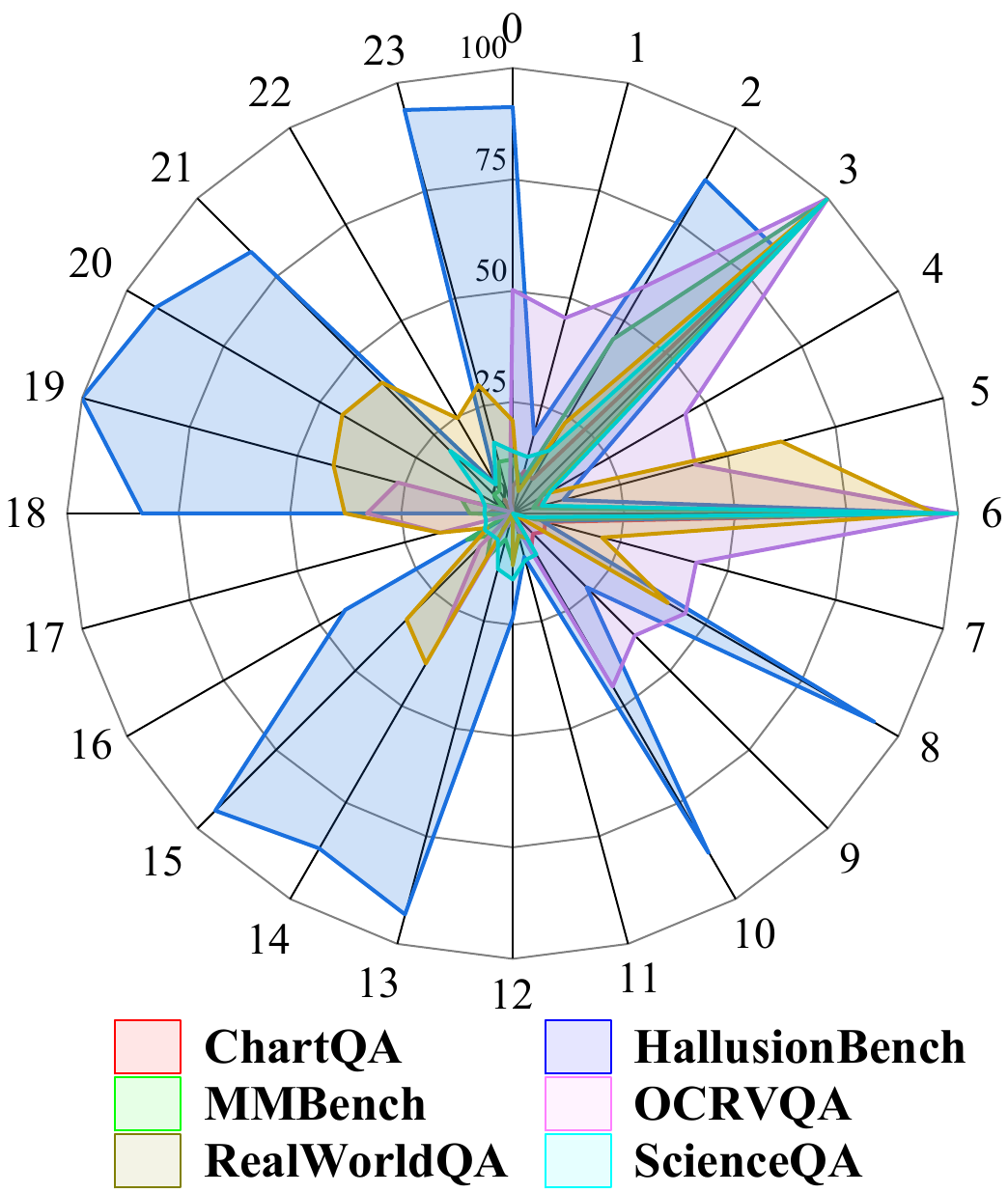}{e}{LoRA $r=32$}\hfill
\ijcnnradarcell{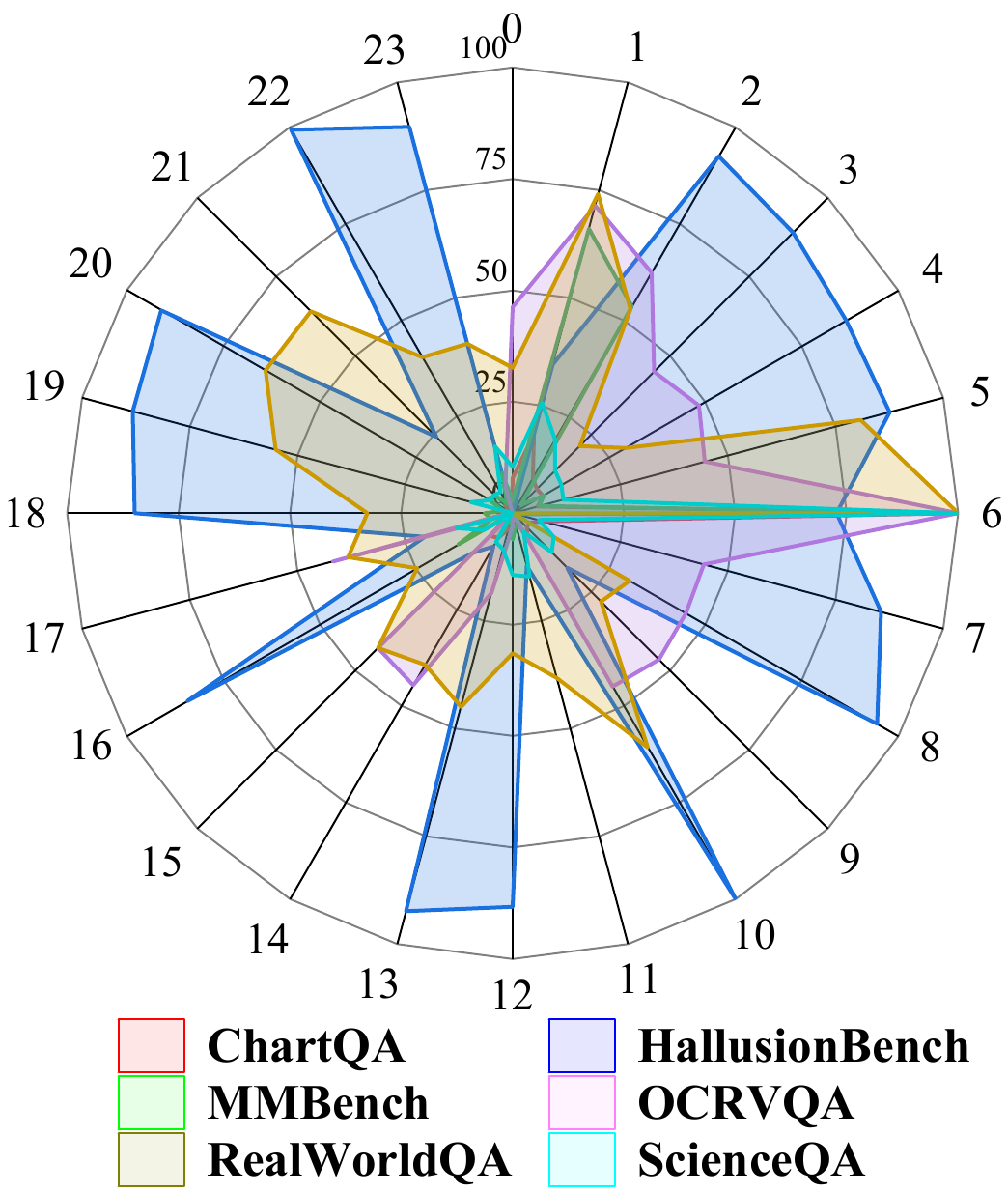}{f}{LoRA $r=128$}\hfill
\ijcnnradarcell{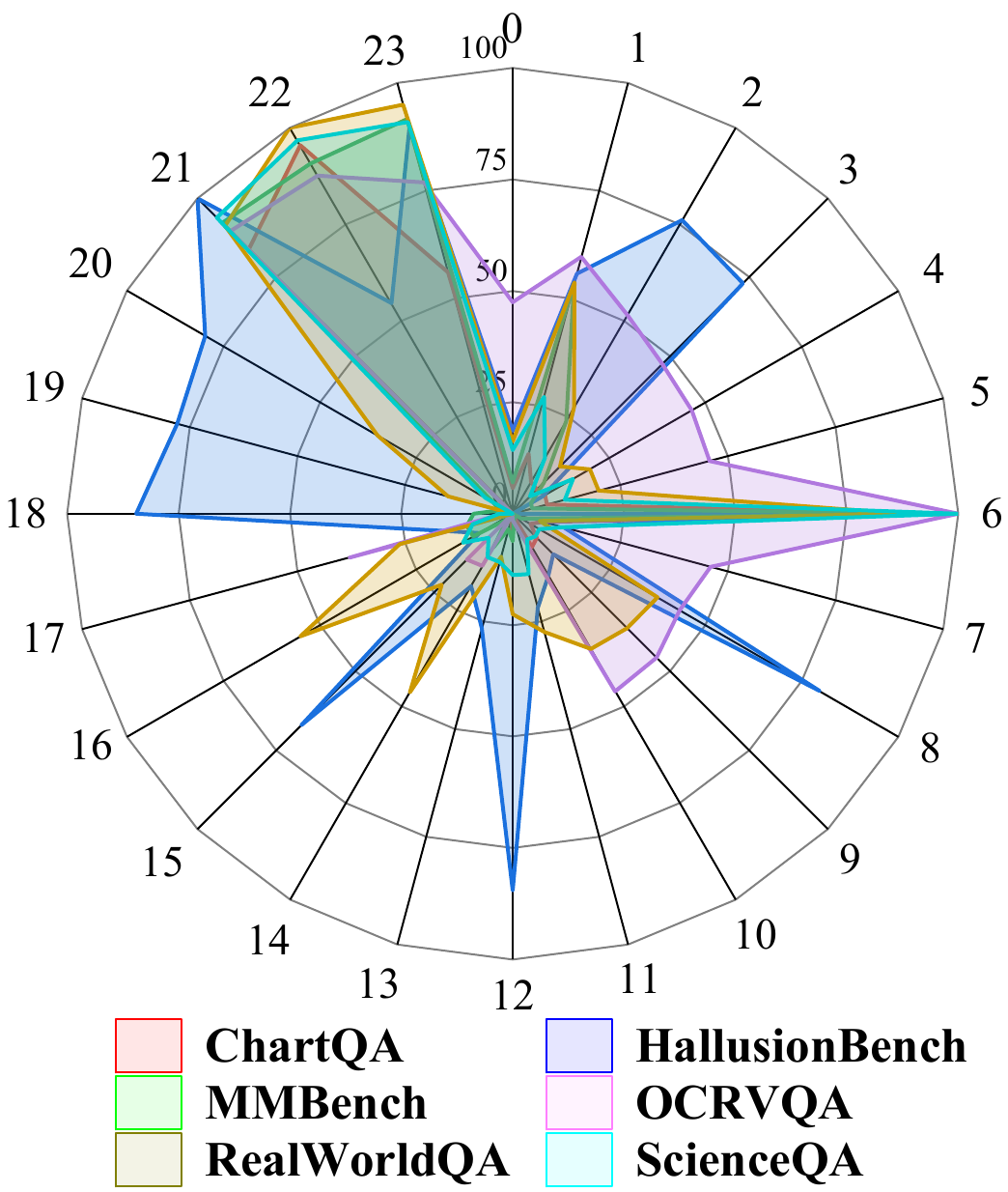}{g}{LoRA $r=512$}\hfill
\ijcnnradarcell{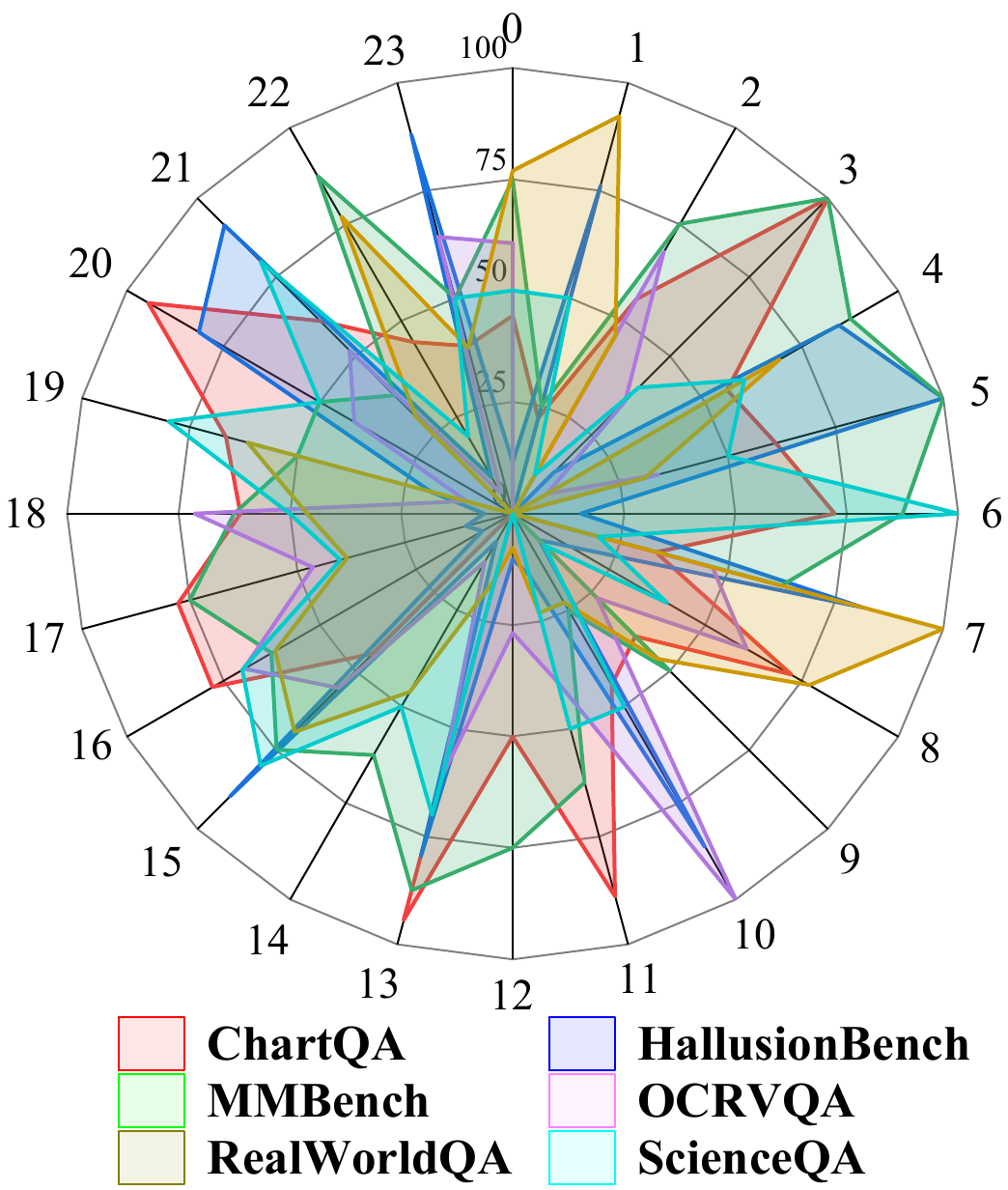}{h}{VeRA}
\vspace{-0.7ex}
\caption{Benchmark scores for all-layer tuning (a) and per-layer PEFT (b--h): Partial-Tuning, IA3, LayerNorm-Tuning, LoRA ($r=32/128/512$), and VeRA.}
\label{fig:main:radar}
\end{figure*}

Figure \ref{fig:main:radar} and Table \ref{tab:main:mmstar_layer_peft} show that the best PEFT gains cluster in early and late layers, broadly overlapping with perturbation-robust, high-statistics layers.

\paragraph{Strong same-budget baselines.}
Under the same adaptation budget ($k=5$), budget-matched heuristics yield only marginal and unstable gains. On InternVL2-1B, first-$k$ LoRA reaches 43.70\% versus the 43.63\% base, while random subsets and single-component tuning are weaker; on Qwen2.5-VL-3B, middle-$k$ LoRA reaches 55.24\% versus the 54.90\% base, whereas attention-only and MLP-only tuning drop to 54.70\% and 54.50\%.

% 说明：将 Table III 从 C 小节标题前移到 D 小节后半段，
% 让前面的正文先铺满版面，再由浮动体自然落到更靠后的位置，减少中后页空白。
\begin{table*}[!t]
\caption{Layer-wise perturbation on MMStar (a) and six benchmarks (b) for~\mbox{InternVL2-1B}. Best in bold; worst underlined.}
\label{tab:main:mmstar_layer_noise}
\centering
\setlength{\belowcaptionskip}{0pt}
\scriptsize
\begingroup
% 说明：左右两张表的“自然宽度”不同，如果各自用 width=\linewidth 缩放，
% 会得到不同缩放因子，导致视觉字号/行高不同，从而出现左右高度不一致。
% 这里先测量两张表的自然宽度，用同一个缩放因子（以较宽者为基准）同时缩放两侧。
\setbox0=\hbox{{\setlength{\tabcolsep}{0.6mm}\renewcommand{\arraystretch}{0.78}\setlength{\extrarowheight}{0.45pt}% InternVL2-1B：MMStar 逐层扰动表（仅 tabular，供 experiments 中并排双栏表复用）
\begin{tabular}[t]{ccccccccccccccc}
\toprule
\multirow{2}{*}{\textbf{L}} & \multicolumn{7}{c}{\rule{0pt}{1.55ex}ActNoise} & \multicolumn{7}{c}{\rule{0pt}{1.55ex}PathNoise} \\
\cmidrule(lr){2-8}\cmidrule(lr){9-15}
\rule{0pt}{1.55ex} & Overall & CP & FGP & IR & LR & MA & S\&T & Overall & CP & FGP & IR & LR & MA & S\&T \\
\midrule
\multicolumn{1}{c|}{0}&\multicolumn{1}{c|}{\textbf{45.6$_{\downarrow\text{0.1}}$}} & 60.0$_{\downarrow\text{1.2}}$ & 36.4$_{\uparrow\text{0.8}}$ & \textbf{53.6$_{\text{0.0}}$} & \textbf{37.2$_{\uparrow\text{2.0}}$} & 47.6$_{\downarrow\text{0.4}}$ & \multicolumn{1}{c|}{38.8$_{\downarrow\text{1.6}}$} &\multicolumn{1}{c|}{44.5$_{\downarrow\text{1.1}}$} & 61.6$_{\uparrow\text{0.4}}$ & 34.0$_{\downarrow\text{1.6}}$ & 51.6$_{\downarrow\text{2.0}}$ & 33.6$_{\downarrow\text{1.6}}$ & 46.4$_{\downarrow\text{1.6}}$ & 40.0$_{\downarrow\text{0.4}}$ \\ 
\multicolumn{1}{c|}{1}&\multicolumn{1}{c|}{45.0$_{\downarrow\text{0.7}}$} & 60.0$_{\downarrow\text{1.2}}$ & 36.0$_{\uparrow\text{0.4}}$ & \textbf{53.6$_{\text{0.0}}$} & 34.0$_{\downarrow\text{1.2}}$ & 45.6$_{\downarrow\text{2.4}}$ & \multicolumn{1}{c|}{40.8$_{\uparrow\text{0.4}}$} &\multicolumn{1}{c|}{44.2$_{\downarrow\text{1.5}}$} & 57.6$_{\downarrow\text{3.6}}$ & 36.0$_{\uparrow\text{0.4}}$ & 51.6$_{\downarrow\text{2.0}}$ & 35.2$_{\text{0.0}}$ & 46.4$_{\downarrow\text{1.6}}$ & 38.4$_{\downarrow\text{2.0}}$ \\ 
\multicolumn{1}{c|}{2}&\multicolumn{1}{c|}{44.9$_{\downarrow\text{0.7}}$} & 60.0$_{\downarrow\text{1.2}}$ & 33.6$_{\downarrow\text{2.0}}$ & 53.2$_{\downarrow\text{0.4}}$ & 34.8$_{\downarrow\text{0.4}}$ & 47.6$_{\downarrow\text{0.4}}$ & \multicolumn{1}{c|}{40.4$_{\text{0.0}}$} &\multicolumn{1}{c|}{\textbf{45.3$_{\downarrow\text{0.4}}$}} & 61.2$_{\text{0.0}}$ & 36.4$_{\uparrow\text{0.8}}$ & 51.6$_{\downarrow\text{2.0}}$ & 36.8$_{\uparrow\text{1.6}}$ & 47.2$_{\downarrow\text{0.8}}$ & 38.4$_{\downarrow\text{2.0}}$ \\ 
\multicolumn{1}{c|}{3}&\multicolumn{1}{c|}{45.5$_{\downarrow\text{0.1}}$} & 62.0$_{\uparrow\text{0.8}}$ & 35.2$_{\downarrow\text{0.4}}$ & 51.6$_{\downarrow\text{2.0}}$ & 34.0$_{\downarrow\text{1.2}}$ & \textbf{49.6$_{\uparrow\text{1.6}}$} & \multicolumn{1}{c|}{40.8$_{\uparrow\text{0.4}}$} &\multicolumn{1}{c|}{43.8$_{\downarrow\text{1.9}}$} & 58.8$_{\downarrow\text{2.4}}$ & 33.2$_{\downarrow\text{2.4}}$ & 51.2$_{\downarrow\text{2.4}}$ & 36.0$_{\uparrow\text{0.8}}$ & 44.4$_{\downarrow\text{3.6}}$ & 39.2$_{\downarrow\text{1.2}}$ \\ 
\multicolumn{1}{c|}{4}&\multicolumn{1}{c|}{45.1$_{\downarrow\text{0.5}}$} & 60.4$_{\downarrow\text{0.8}}$ & 36.0$_{\uparrow\text{0.4}}$ & 53.2$_{\downarrow\text{0.4}}$ & 33.2$_{\downarrow\text{2.0}}$ & 49.2$_{\uparrow\text{1.2}}$ & \multicolumn{1}{c|}{38.8$_{\downarrow\text{1.6}}$} &\multicolumn{1}{c|}{44.9$_{\downarrow\text{0.8}}$} & 59.6$_{\downarrow\text{1.6}}$ & 36.4$_{\uparrow\text{0.8}}$ & 51.2$_{\downarrow\text{2.4}}$ & 36.0$_{\uparrow\text{0.8}}$ & 47.2$_{\downarrow\text{0.8}}$ & 38.8$_{\downarrow\text{1.6}}$ \\ 
\multicolumn{1}{c|}{5}&\multicolumn{1}{c|}{45.4$_{\downarrow\text{0.3}}$} & 60.0$_{\downarrow\text{1.2}}$ & 36.4$_{\uparrow\text{0.8}}$ & 52.4$_{\downarrow\text{1.2}}$ & 35.6$_{\uparrow\text{0.4}}$ & 47.2$_{\downarrow\text{0.8}}$ & \multicolumn{1}{c|}{40.8$_{\uparrow\text{0.4}}$} &\multicolumn{1}{c|}{44.5$_{\downarrow\text{1.2}}$} & 61.2$_{\text{0.0}}$ & 35.6$_{\text{0.0}}$ & 51.2$_{\downarrow\text{2.4}}$ & \textbf{37.2$_{\uparrow\text{2.0}}$} & 45.6$_{\downarrow\text{2.4}}$ & 36.0$_{\downarrow\text{4.4}}$ \\ 
\multicolumn{1}{c|}{6}&\multicolumn{1}{c|}{44.6$_{\downarrow\text{1.1}}$} & 60.4$_{\downarrow\text{0.8}}$ & 36.0$_{\uparrow\text{0.4}}$ & 51.2$_{\downarrow\text{2.4}}$ & 36.0$_{\uparrow\text{0.8}}$ & 45.6$_{\downarrow\text{2.4}}$ & \multicolumn{1}{c|}{38.4$_{\downarrow\text{2.0}}$} &\multicolumn{1}{c|}{\underline{42.7$_{\downarrow\text{3.0}}$}} & 57.2$_{\downarrow\text{4.0}}$ & 33.2$_{\downarrow\text{2.4}}$ & 51.2$_{\downarrow\text{2.4}}$ & \underline{33.2$_{\downarrow\text{2.0}}$} & 46.8$_{\downarrow\text{1.2}}$ & \underline{34.4$_{\downarrow\text{6.0}}$} \\ 
\multicolumn{1}{c|}{7}&\multicolumn{1}{c|}{44.7$_{\downarrow\text{0.9}}$} & 60.0$_{\downarrow\text{1.2}}$ & 33.6$_{\downarrow\text{2.0}}$ & 53.2$_{\downarrow\text{0.4}}$ & 34.8$_{\downarrow\text{0.4}}$ & 47.2$_{\downarrow\text{0.8}}$ & \multicolumn{1}{c|}{39.6$_{\downarrow\text{0.8}}$} &\multicolumn{1}{c|}{\textbf{45.3$_{\downarrow\text{0.4}}$}} & 60.0$_{\downarrow\text{1.2}}$ & 35.6$_{\text{0.0}}$ & 51.2$_{\downarrow\text{2.4}}$ & 36.4$_{\uparrow\text{1.2}}$ & 46.4$_{\downarrow\text{1.6}}$ & \textbf{42.0$_{\uparrow\text{1.6}}$} \\ \midrule
\multicolumn{1}{c|}{8}&\multicolumn{1}{c|}{44.8$_{\downarrow\text{0.9}}$} & 60.4$_{\downarrow\text{0.8}}$ & 36.0$_{\uparrow\text{0.4}}$ & 52.0$_{\downarrow\text{1.6}}$ & 34.4$_{\downarrow\text{0.8}}$ & 46.4$_{\downarrow\text{1.6}}$ & \multicolumn{1}{c|}{39.6$_{\downarrow\text{0.8}}$} &\multicolumn{1}{c|}{44.6$_{\downarrow\text{1.1}}$} & 58.0$_{\downarrow\text{3.2}}$ & \textbf{37.2$_{\uparrow\text{1.6}}$} & 53.6$_{\text{0.0}}$ & 34.8$_{\downarrow\text{0.4}}$ & 46.8$_{\downarrow\text{1.2}}$ & 37.2$_{\downarrow\text{3.2}}$ \\ 
\multicolumn{1}{c|}{9}&\multicolumn{1}{c|}{44.5$_{\downarrow\text{1.1}}$} & 59.6$_{\downarrow\text{1.6}}$ & 35.2$_{\downarrow\text{0.4}}$ & 52.0$_{\downarrow\text{1.6}}$ & 36.0$_{\uparrow\text{0.8}}$ & \underline{45.2$_{\downarrow\text{2.8}}$} & \multicolumn{1}{c|}{39.2$_{\downarrow\text{1.2}}$} &\multicolumn{1}{c|}{44.6$_{\downarrow\text{1.1}}$} & 58.0$_{\downarrow\text{3.2}}$ & \textbf{37.2$_{\uparrow\text{1.6}}$} & 53.6$_{\text{0.0}}$ & 34.8$_{\downarrow\text{0.4}}$ & 46.8$_{\downarrow\text{1.2}}$ & 37.2$_{\downarrow\text{3.2}}$ \\ 
\multicolumn{1}{c|}{10}&\multicolumn{1}{c|}{44.7$_{\downarrow\text{1.0}}$} & 60.4$_{\downarrow\text{0.8}}$ & 33.2$_{\downarrow\text{2.4}}$ & 52.8$_{\downarrow\text{0.8}}$ & 35.6$_{\uparrow\text{0.4}}$ & 46.0$_{\downarrow\text{2.0}}$ & \multicolumn{1}{c|}{40.0$_{\downarrow\text{0.4}}$} &\multicolumn{1}{c|}{44.5$_{\downarrow\text{1.1}}$} & 59.6$_{\downarrow\text{1.6}}$ & 35.6$_{\text{0.0}}$ & 52.0$_{\downarrow\text{1.6}}$ & 34.8$_{\downarrow\text{0.4}}$ & 45.2$_{\downarrow\text{2.8}}$ & 40.0$_{\downarrow\text{0.4}}$ \\ 
\multicolumn{1}{c|}{11}&\multicolumn{1}{c|}{45.1$_{\downarrow\text{0.6}}$} & \textbf{62.4$_{\uparrow\text{1.2}}$} & 36.8$_{\uparrow\text{1.2}}$ & 52.4$_{\downarrow\text{1.2}}$ & 34.8$_{\downarrow\text{0.4}}$ & \underline{45.2$_{\downarrow\text{2.8}}$} & \multicolumn{1}{c|}{38.8$_{\downarrow\text{1.6}}$} &\multicolumn{1}{c|}{44.5$_{\downarrow\text{1.1}}$} & 59.6$_{\downarrow\text{1.6}}$ & 35.6$_{\text{0.0}}$ & 52.0$_{\downarrow\text{1.6}}$ & 34.8$_{\downarrow\text{0.4}}$ & 45.2$_{\downarrow\text{2.8}}$ & 40.0$_{\downarrow\text{0.4}}$ \\ 
\multicolumn{1}{c|}{12}&\multicolumn{1}{c|}{\underline{43.9$_{\downarrow\text{1.8}}$}} & 59.2$_{\downarrow\text{2.0}}$ & 32.4$_{\downarrow\text{3.2}}$ & \underline{50.4$_{\downarrow\text{3.2}}$} & 34.8$_{\downarrow\text{0.4}}$ & 47.2$_{\downarrow\text{0.8}}$ & \multicolumn{1}{c|}{39.2$_{\downarrow\text{1.2}}$} &\multicolumn{1}{c|}{44.1$_{\downarrow\text{1.5}}$} & 60.4$_{\downarrow\text{0.8}}$ & 35.6$_{\text{0.0}}$ & 51.2$_{\downarrow\text{2.4}}$ & 35.6$_{\uparrow\text{0.4}}$ & 47.2$_{\downarrow\text{0.8}}$ & 34.8$_{\downarrow\text{5.6}}$ \\ 
\multicolumn{1}{c|}{13}&\multicolumn{1}{c|}{44.0$_{\downarrow\text{1.7}}$} & 59.6$_{\downarrow\text{1.6}}$ & 32.4$_{\downarrow\text{3.2}}$ & 52.8$_{\downarrow\text{0.8}}$ & 34.4$_{\downarrow\text{0.8}}$ & 47.6$_{\downarrow\text{0.4}}$ & \multicolumn{1}{c|}{\underline{37.2$_{\downarrow\text{3.2}}$}} &\multicolumn{1}{c|}{\underline{42.7$_{\downarrow\text{3.0}}$}} & \underline{56.4$_{\downarrow\text{4.8}}$} & 32.8$_{\downarrow\text{2.8}}$ & \underline{49.6$_{\downarrow\text{4.0}}$} & 34.0$_{\downarrow\text{1.2}}$ & 46.0$_{\downarrow\text{2.0}}$ & 37.2$_{\downarrow\text{3.2}}$ \\ 
\multicolumn{1}{c|}{14}&\multicolumn{1}{c|}{44.0$_{\downarrow\text{1.7}}$} & 60.4$_{\downarrow\text{0.8}}$ & 32.4$_{\downarrow\text{3.2}}$ & 50.8$_{\downarrow\text{2.8}}$ & 34.4$_{\downarrow\text{0.8}}$ & 48.4$_{\uparrow\text{0.4}}$ & \multicolumn{1}{c|}{37.6$_{\downarrow\text{2.8}}$} &\multicolumn{1}{c|}{43.9$_{\downarrow\text{1.7}}$} & 59.6$_{\downarrow\text{1.6}}$ & \underline{31.2$_{\downarrow\text{4.4}}$} & 53.2$_{\downarrow\text{0.4}}$ & 34.8$_{\downarrow\text{0.4}}$ & 47.2$_{\downarrow\text{0.8}}$ & 37.6$_{\downarrow\text{2.8}}$ \\ 
\multicolumn{1}{c|}{15}&\multicolumn{1}{c|}{44.4$_{\downarrow\text{1.3}}$} & \underline{57.2$_{\downarrow\text{4.0}}$} & 33.6$_{\downarrow\text{2.0}}$ & \textbf{53.6$_{\text{0.0}}$} & 36.8$_{\uparrow\text{1.6}}$ & 47.2$_{\downarrow\text{0.8}}$ & \multicolumn{1}{c|}{38.0$_{\downarrow\text{2.4}}$} &\multicolumn{1}{c|}{44.5$_{\downarrow\text{1.1}}$} & 59.2$_{\downarrow\text{2.0}}$ & 34.4$_{\downarrow\text{1.2}}$ & 52.0$_{\downarrow\text{1.6}}$ & 35.2$_{\text{0.0}}$ & 47.2$_{\downarrow\text{0.8}}$ & 39.2$_{\downarrow\text{1.2}}$ \\ \midrule
\multicolumn{1}{c|}{16}&\multicolumn{1}{c|}{44.5$_{\downarrow\text{1.2}}$} & 60.4$_{\downarrow\text{0.8}}$ & 34.4$_{\downarrow\text{1.2}}$ & \underline{50.4$_{\downarrow\text{3.2}}$} & 35.6$_{\uparrow\text{0.4}}$ & 46.8$_{\downarrow\text{1.2}}$ & \multicolumn{1}{c|}{39.2$_{\downarrow\text{1.2}}$} &\multicolumn{1}{c|}{44.4$_{\downarrow\text{1.3}}$} & 58.8$_{\downarrow\text{2.4}}$ & 31.6$_{\downarrow\text{4.0}}$ & 52.8$_{\downarrow\text{0.8}}$ & 36.4$_{\uparrow\text{1.2}}$ & \textbf{48.0$_{\text{0.0}}$} & 38.8$_{\downarrow\text{1.6}}$ \\ 
\multicolumn{1}{c|}{17}&\multicolumn{1}{c|}{44.1$_{\downarrow\text{1.6}}$} & 59.6$_{\downarrow\text{1.6}}$ & \underline{32.0$_{\downarrow\text{3.6}}$} & 52.0$_{\downarrow\text{1.6}}$ & 36.4$_{\uparrow\text{1.2}}$ & \underline{45.2$_{\downarrow\text{2.8}}$} & \multicolumn{1}{c|}{39.2$_{\downarrow\text{1.2}}$} &\multicolumn{1}{c|}{44.1$_{\downarrow\text{1.5}}$} & 56.8$_{\downarrow\text{4.4}}$ & 34.4$_{\downarrow\text{1.2}}$ & 51.2$_{\downarrow\text{2.4}}$ & 35.6$_{\uparrow\text{0.4}}$ & 46.0$_{\downarrow\text{2.0}}$ & 40.8$_{\uparrow\text{0.4}}$ \\ 
\multicolumn{1}{c|}{18}&\multicolumn{1}{c|}{45.3$_{\downarrow\text{0.3}}$} & 61.6$_{\uparrow\text{0.4}}$ & 34.8$_{\downarrow\text{0.8}}$ & 52.4$_{\downarrow\text{1.2}}$ & 36.4$_{\uparrow\text{1.2}}$ & 46.8$_{\downarrow\text{1.2}}$ & \multicolumn{1}{c|}{40.0$_{\downarrow\text{0.4}}$} &\multicolumn{1}{c|}{44.7$_{\downarrow\text{1.0}}$} & 60.0$_{\downarrow\text{1.2}}$ & 33.2$_{\downarrow\text{2.4}}$ & 52.8$_{\downarrow\text{0.8}}$ & 35.2$_{\text{0.0}}$ & 46.0$_{\downarrow\text{2.0}}$ & 40.8$_{\uparrow\text{0.4}}$ \\ 
\multicolumn{1}{c|}{19}&\multicolumn{1}{c|}{45.0$_{\downarrow\text{0.7}}$} & 61.2$_{\text{0.0}}$ & 35.2$_{\downarrow\text{0.4}}$ & \textbf{53.6$_{\text{0.0}}$} & 34.0$_{\downarrow\text{1.2}}$ & 46.0$_{\downarrow\text{2.0}}$ & \multicolumn{1}{c|}{40.0$_{\downarrow\text{0.4}}$} &\multicolumn{1}{c|}{45.0$_{\downarrow\text{0.7}}$} & 60.0$_{\downarrow\text{1.2}}$ & 35.6$_{\text{0.0}}$ & 52.4$_{\downarrow\text{1.2}}$ & 35.6$_{\uparrow\text{0.4}}$ & 45.6$_{\downarrow\text{2.4}}$ & 40.8$_{\uparrow\text{0.4}}$ \\ 
\multicolumn{1}{c|}{20}&\multicolumn{1}{c|}{44.8$_{\downarrow\text{0.9}}$} & 62.0$_{\uparrow\text{0.8}}$ & 35.2$_{\downarrow\text{0.4}}$ & 52.0$_{\downarrow\text{1.6}}$ & 34.0$_{\downarrow\text{1.2}}$ & 47.6$_{\downarrow\text{0.4}}$ & \multicolumn{1}{c|}{38.0$_{\downarrow\text{2.4}}$} &\multicolumn{1}{c|}{44.9$_{\downarrow\text{0.8}}$} & 60.4$_{\downarrow\text{0.8}}$ & 36.0$_{\uparrow\text{0.4}}$ & 51.6$_{\downarrow\text{2.0}}$ & 36.0$_{\uparrow\text{0.8}}$ & 44.8$_{\downarrow\text{3.2}}$ & 40.4$_{\text{0.0}}$ \\ 
\multicolumn{1}{c|}{21}&\multicolumn{1}{c|}{44.6$_{\downarrow\text{1.1}}$} & 60.0$_{\downarrow\text{1.2}}$ & 33.2$_{\downarrow\text{2.4}}$ & 51.6$_{\downarrow\text{2.0}}$ & 35.6$_{\uparrow\text{0.4}}$ & 48.4$_{\uparrow\text{0.4}}$ & \multicolumn{1}{c|}{38.8$_{\downarrow\text{1.6}}$} &\multicolumn{1}{c|}{44.1$_{\downarrow\text{1.6}}$} & 59.6$_{\downarrow\text{1.6}}$ & 33.2$_{\downarrow\text{2.4}}$ & 52.8$_{\downarrow\text{0.8}}$ & 35.6$_{\uparrow\text{0.4}}$ & 44.8$_{\downarrow\text{3.2}}$ & 38.4$_{\downarrow\text{2.0}}$ \\ 
\multicolumn{1}{c|}{22}&\multicolumn{1}{c|}{44.9$_{\downarrow\text{0.7}}$} & 60.8$_{\downarrow\text{0.4}}$ & 36.0$_{\uparrow\text{0.4}}$ & 51.2$_{\downarrow\text{2.4}}$ & \underline{32.8$_{\downarrow\text{2.4}}$} & 47.6$_{\downarrow\text{0.4}}$ & \multicolumn{1}{c|}{\textbf{41.2$_{\uparrow\text{0.8}}$}} &\multicolumn{1}{c|}{44.5$_{\downarrow\text{1.1}}$} & \textbf{62.0$_{\uparrow\text{0.8}}$} & 35.2$_{\downarrow\text{0.4}}$ & \textbf{54.0$_{\uparrow\text{0.4}}$} & 36.4$_{\uparrow\text{1.2}}$ & \underline{42.0$_{\downarrow\text{6.0}}$} & 37.6$_{\downarrow\text{2.8}}$ \\ 
\multicolumn{1}{c|}{23}&\multicolumn{1}{c|}{44.7$_{\downarrow\text{1.0}}$} & 60.4$_{\downarrow\text{0.8}}$ & \textbf{37.6$_{\uparrow\text{2.0}}$} & 51.2$_{\downarrow\text{2.4}}$ & 35.6$_{\uparrow\text{0.4}}$ & 46.0$_{\downarrow\text{2.0}}$ & \multicolumn{1}{c|}{\underline{37.2$_{\downarrow\text{3.2}}$}} &\multicolumn{1}{c|}{44.1$_{\downarrow\text{1.5}}$} & 59.6$_{\downarrow\text{1.6}}$ & 33.6$_{\downarrow\text{2.0}}$ & 51.2$_{\downarrow\text{2.4}}$ & \underline{33.2$_{\downarrow\text{2.0}}$} & 46.8$_{\downarrow\text{1.2}}$ & 40.4$_{\text{0.0}}$ \\ 
\bottomrule
\end{tabular}
}}
\setbox2=\hbox{{\setlength{\tabcolsep}{0.6mm}\renewcommand{\arraystretch}{0.78}\setlength{\extrarowheight}{0.45pt}% InternVL2-1B：Benchmarks 逐层扰动表（仅 tabular，供 experiments 中并排双栏表复用）
\begin{tabular}[t]{@{}ccccccccccccc@{}}
\toprule
\multirow{2}{*}{\textbf{Layer}} & \multicolumn{6}{c}{\rule{0pt}{1.55ex}ActNoise} & \multicolumn{6}{c}{\rule{0pt}{1.55ex}PathNoise} \\
\cmidrule(lr){2-7}\cmidrule(lr){8-13}
\rule{0pt}{1.55ex} & CQA & HB & MMB & OCRVQA & RWQA & \multicolumn{1}{c|}{SQA} & CQA & HB & MMB & OCRVQA & RWQA & SQA \\ \midrule
\multicolumn{1}{c|}{0}&55.8$_{\downarrow\text{2.0}}$ & 31.5$_{\downarrow\text{3.7}}$ & 67.4$_{\uparrow\text{0.3}}$ & 43.3$_{\downarrow\text{0.2}}$ & 48.4$_{\downarrow\text{0.4}}$ & \multicolumn{1}{c|}{87.5$_{\text{0.0}}$} &53.7$_{\downarrow\text{4.2}}$ & 32.3$_{\downarrow\text{2.9}}$ & 67.0$_{\downarrow\text{0.2}}$ & 42.8$_{\downarrow\text{0.7}}$ & 48.1$_{\downarrow\text{0.7}}$ & \textbf{87.8$_{\uparrow\text{0.3}}$} \\ 
\multicolumn{1}{c|}{1}&57.4$_{\downarrow\text{0.5}}$ & 31.8$_{\downarrow\text{3.4}}$ & 67.7$_{\uparrow\text{0.5}}$ & \textbf{43.5$_{\text{0.0}}$} & 48.5$_{\downarrow\text{0.3}}$ & \multicolumn{1}{c|}{87.6$_{\uparrow\text{0.1}}$} &54.0$_{\downarrow\text{3.9}}$ & 33.2$_{\downarrow\text{2.0}}$ & 65.6$_{\downarrow\text{1.5}}$ & 42.7$_{\downarrow\text{0.8}}$ & 47.2$_{\downarrow\text{1.6}}$ & 86.9$_{\downarrow\text{0.5}}$ \\ 
\multicolumn{1}{c|}{2}&58.2$_{\uparrow\text{0.3}}$ & 32.4$_{\downarrow\text{2.8}}$ & 67.8$_{\uparrow\text{0.6}}$ & \textbf{43.5$_{\text{0.0}}$} & 48.4$_{\downarrow\text{0.4}}$ & \multicolumn{1}{c|}{87.6$_{\uparrow\text{0.1}}$} &54.3$_{\downarrow\text{3.6}}$ & 32.9$_{\downarrow\text{2.3}}$ & 67.0$_{\downarrow\text{0.2}}$ & 42.8$_{\downarrow\text{0.7}}$ & 46.0$_{\downarrow\text{2.7}}$ & 86.7$_{\downarrow\text{0.7}}$ \\ 
\multicolumn{1}{c|}{3}&55.5$_{\downarrow\text{2.4}}$ & 35.6$_{\uparrow\text{0.4}}$ & 67.7$_{\uparrow\text{0.5}}$ & 43.3$_{\downarrow\text{0.2}}$ & 47.3$_{\downarrow\text{1.4}}$ & \multicolumn{1}{c|}{87.6$_{\uparrow\text{0.1}}$} &47.5$_{\downarrow\text{10.4}}$ & 34.2$_{\downarrow\text{1.0}}$ & 65.4$_{\downarrow\text{1.8}}$ & 42.3$_{\downarrow\text{1.2}}$ & 45.2$_{\downarrow\text{3.5}}$ & 87.0$_{\downarrow\text{0.4}}$ \\ 
\multicolumn{1}{c|}{4}&57.1$_{\downarrow\text{0.8}}$ & 31.9$_{\downarrow\text{3.3}}$ & \underline{66.7$_{\downarrow\text{0.5}}$} & 43.4$_{\downarrow\text{0.1}}$ & 48.2$_{\downarrow\text{0.5}}$ & \multicolumn{1}{c|}{87.5$_{\text{0.0}}$} &54.9$_{\downarrow\text{3.0}}$ & 31.8$_{\downarrow\text{3.4}}$ & 67.3$_{\uparrow\text{0.1}}$ & 43.0$_{\downarrow\text{0.5}}$ & 48.4$_{\downarrow\text{0.4}}$ & 87.1$_{\downarrow\text{0.3}}$ \\ 
\multicolumn{1}{c|}{5}&56.6$_{\downarrow\text{1.3}}$ & \textbf{35.7$_{\uparrow\text{0.5}}$} & 67.3$_{\uparrow\text{0.1}}$ & 43.3$_{\downarrow\text{0.2}}$ & 47.5$_{\downarrow\text{1.3}}$ & \multicolumn{1}{c|}{87.5$_{\uparrow\text{0.1}}$} &55.0$_{\downarrow\text{2.8}}$ & 34.3$_{\downarrow\text{1.0}}$ & \textbf{67.6$_{\uparrow\text{0.4}}$} & 43.1$_{\downarrow\text{0.4}}$ & 45.9$_{\downarrow\text{2.9}}$ & 87.5$_{\uparrow\text{0.1}}$ \\ 
\multicolumn{1}{c|}{6}&52.8$_{\downarrow\text{5.0}}$ & 33.4$_{\downarrow\text{1.8}}$ & 67.6$_{\uparrow\text{0.4}}$ & 43.0$_{\downarrow\text{0.5}}$ & 48.1$_{\downarrow\text{0.7}}$ & \multicolumn{1}{c|}{87.5$_{\uparrow\text{0.1}}$} &50.2$_{\downarrow\text{7.6}}$ & 32.0$_{\downarrow\text{3.2}}$ & 65.0$_{\downarrow\text{2.1}}$ & 41.8$_{\downarrow\text{1.7}}$ & 45.4$_{\downarrow\text{3.4}}$ & \underline{85.9$_{\downarrow\text{1.5}}$} \\ 
\multicolumn{1}{c|}{7}&57.8$_{\downarrow\text{0.1}}$ & 35.3$_{\uparrow\text{0.1}}$ & 67.2$_{\text{0.0}}$ & 43.4$_{\downarrow\text{0.2}}$ & 47.3$_{\downarrow\text{1.4}}$ & \multicolumn{1}{c|}{87.7$_{\uparrow\text{0.2}}$} &56.8$_{\downarrow\text{1.1}}$ & 31.5$_{\downarrow\text{3.7}}$ & 67.4$_{\uparrow\text{0.3}}$ & 43.3$_{\downarrow\text{0.2}}$ & 46.1$_{\downarrow\text{2.6}}$ & 87.3$_{\downarrow\text{0.1}}$ \\ \midrule
\multicolumn{1}{c|}{8}&56.9$_{\downarrow\text{1.0}}$ & 32.5$_{\downarrow\text{2.7}}$ & \underline{66.7$_{\downarrow\text{0.5}}$} & 43.2$_{\downarrow\text{0.3}}$ & 45.2$_{\downarrow\text{3.5}}$ & \multicolumn{1}{c|}{\textbf{88.1$_{\uparrow\text{0.6}}$}} &55.7$_{\downarrow\text{2.2}}$ & 31.2$_{\downarrow\text{4.0}}$ & 67.2$_{\text{0.0}}$ & 43.4$_{\downarrow\text{0.1}}$ & \underline{43.0$_{\downarrow\text{7.2}}$} & 87.5$_{\text{0.0}}$ \\ 
\multicolumn{1}{c|}{9}&56.6$_{\downarrow\text{1.3}}$ & 35.1$_{\downarrow\text{0.1}}$ & 67.0$_{\downarrow\text{0.2}}$ & 43.1$_{\downarrow\text{0.4}}$ & 47.8$_{\downarrow\text{0.9}}$ & \multicolumn{1}{c|}{87.9$_{\uparrow\text{0.4}}$} &55.3$_{\downarrow\text{2.7}}$ & 32.7$_{\downarrow\text{2.5}}$ & 67.3$_{\uparrow\text{0.1}}$ & 43.2$_{\downarrow\text{0.3}}$ & 46.5$_{\downarrow\text{2.2}}$ & 87.8$_{\uparrow\text{0.3}}$ \\ 
\multicolumn{1}{c|}{10}&56.5$_{\downarrow\text{1.4}}$ & 31.6$_{\downarrow\text{3.7}}$ & 67.5$_{\uparrow\text{0.3}}$ & 43.3$_{\downarrow\text{0.2}}$ & 45.6$_{\downarrow\text{3.1}}$ & \multicolumn{1}{c|}{87.9$_{\uparrow\text{0.4}}$} &55.1$_{\downarrow\text{2.8}}$ & 34.7$_{\downarrow\text{0.5}}$ & 67.4$_{\uparrow\text{0.3}}$ & 43.3$_{\downarrow\text{0.2}}$ & 47.0$_{\downarrow\text{1.7}}$ & 87.7$_{\uparrow\text{0.2}}$ \\ 
\multicolumn{1}{c|}{11}&56.9$_{\downarrow\text{1.0}}$ & 32.4$_{\downarrow\text{2.8}}$ & 67.7$_{\uparrow\text{0.5}}$ & 43.2$_{\downarrow\text{0.3}}$ & 45.2$_{\downarrow\text{3.6}}$ & \multicolumn{1}{c|}{87.8$_{\uparrow\text{0.3}}$} &55.2$_{\downarrow\text{2.7}}$ & 33.2$_{\downarrow\text{2.0}}$ & 67.6$_{\uparrow\text{0.4}}$ & 43.0$_{\downarrow\text{0.5}}$ & 45.9$_{\downarrow\text{2.8}}$ & 87.6$_{\uparrow\text{0.1}}$ \\ 
\multicolumn{1}{c|}{12}&57.4$_{\downarrow\text{0.5}}$ & 33.3$_{\downarrow\text{1.9}}$ & 67.2$_{\text{0.0}}$ & 43.3$_{\downarrow\text{0.2}}$ & 47.8$_{\downarrow\text{0.9}}$ & \multicolumn{1}{c|}{87.6$_{\uparrow\text{0.1}}$} &54.9$_{\downarrow\text{3.1}}$ & 35.0$_{\downarrow\text{0.3}}$ & 67.2$_{\text{0.0}}$ & 43.3$_{\downarrow\text{0.2}}$ & 47.6$_{\downarrow\text{1.2}}$ & 87.5$_{\text{0.0}}$ \\ 
\multicolumn{1}{c|}{13}&\underline{51.9$_{\downarrow\text{6.0}}$} & 35.5$_{\uparrow\text{0.3}}$ & 67.6$_{\uparrow\text{0.4}}$ & \underline{42.5$_{\downarrow\text{1.0}}$} & 46.9$_{\downarrow\text{1.8}}$ & \multicolumn{1}{c|}{\underline{87.0$_{\downarrow\text{0.4}}$}} &\underline{43.6$_{\downarrow\text{14.2}}$} & 30.6$_{\downarrow\text{4.6}}$ & 65.5$_{\downarrow\text{1.6}}$ & \underline{39.9$_{\downarrow\text{3.6}}$} & 47.2$_{\downarrow\text{1.6}}$ & 87.4$_{\text{0.0}}$ \\ 
\multicolumn{1}{c|}{14}&52.5$_{\downarrow\text{5.4}}$ & 31.4$_{\downarrow\text{3.9}}$ & 67.3$_{\uparrow\text{0.1}}$ & 42.6$_{\downarrow\text{0.9}}$ & \underline{44.3$_{\downarrow\text{4.4}}$} & \multicolumn{1}{c|}{87.4$_{\text{0.0}}$} &55.3$_{\downarrow\text{2.6}}$ & 34.4$_{\downarrow\text{0.8}}$ & 66.5$_{\downarrow\text{0.7}}$ & 43.2$_{\downarrow\text{0.3}}$ & 47.2$_{\downarrow\text{1.6}}$ & 87.5$_{\uparrow\text{0.1}}$ \\ 
\multicolumn{1}{c|}{15}&57.2$_{\downarrow\text{0.7}}$ & 31.2$_{\downarrow\text{4.0}}$ & 67.4$_{\uparrow\text{0.2}}$ & 43.0$_{\downarrow\text{0.6}}$ & 46.9$_{\downarrow\text{1.8}}$ & \multicolumn{1}{c|}{87.5$_{\text{0.0}}$} &56.0$_{\downarrow\text{1.9}}$ & 36.0$_{\uparrow\text{0.8}}$ & 67.4$_{\uparrow\text{0.3}}$ & 42.8$_{\downarrow\text{0.7}}$ & \textbf{49.5$_{\uparrow\text{0.8}}$} & 87.4$_{\text{0.0}}$ \\ \midrule
\multicolumn{1}{c|}{16}&56.9$_{\downarrow\text{1.0}}$ & 35.3$_{\uparrow\text{0.1}}$ & \textbf{67.9$_{\uparrow\text{0.7}}$} & 42.9$_{\downarrow\text{0.6}}$ & 48.5$_{\downarrow\text{0.3}}$ & \multicolumn{1}{c|}{87.5$_{\uparrow\text{0.1}}$} &\textbf{58.0$_{\uparrow\text{0.1}}$} & 31.4$_{\downarrow\text{3.8}}$ & 66.8$_{\downarrow\text{0.4}}$ & 43.1$_{\downarrow\text{0.4}}$ & 47.7$_{\downarrow\text{1.0}}$ & 87.1$_{\downarrow\text{0.3}}$ \\ 
\multicolumn{1}{c|}{17}&58.5$_{\uparrow\text{0.6}}$ & 35.2$_{\text{0.0}}$ & 67.0$_{\downarrow\text{0.2}}$ & 43.3$_{\downarrow\text{0.2}}$ & 48.5$_{\downarrow\text{0.3}}$ & \multicolumn{1}{c|}{\underline{87.0$_{\downarrow\text{0.4}}$}} &57.9$_{\text{0.0}}$ & \textbf{36.2$_{\uparrow\text{1.0}}$} & 66.7$_{\downarrow\text{0.5}}$ & 43.3$_{\downarrow\text{0.2}}$ & 47.1$_{\downarrow\text{1.7}}$ & 87.6$_{\uparrow\text{0.2}}$ \\ 
\multicolumn{1}{c|}{18}&57.5$_{\downarrow\text{0.4}}$ & 34.1$_{\downarrow\text{1.1}}$ & 67.4$_{\uparrow\text{0.2}}$ & 43.3$_{\downarrow\text{0.2}}$ & 47.3$_{\downarrow\text{1.4}}$ & \multicolumn{1}{c|}{87.3$_{\downarrow\text{0.1}}$} &56.9$_{\downarrow\text{1.0}}$ & 31.5$_{\downarrow\text{3.7}}$ & 66.1$_{\downarrow\text{1.1}}$ & 43.2$_{\downarrow\text{0.3}}$ & 47.3$_{\downarrow\text{1.4}}$ & 87.3$_{\downarrow\text{0.1}}$ \\ 
\multicolumn{1}{c|}{19}&57.5$_{\downarrow\text{0.4}}$ & 34.9$_{\downarrow\text{0.3}}$ & 67.0$_{\downarrow\text{0.2}}$ & 43.3$_{\downarrow\text{0.2}}$ & 47.3$_{\downarrow\text{1.4}}$ & \multicolumn{1}{c|}{87.6$_{\uparrow\text{0.1}}$} &55.6$_{\downarrow\text{2.3}}$ & 31.2$_{\downarrow\text{4.0}}$ & 67.2$_{\text{0.0}}$ & 43.4$_{\downarrow\text{0.1}}$ & 46.8$_{\downarrow\text{2.0}}$ & 87.2$_{\downarrow\text{0.2}}$ \\ 
\multicolumn{1}{c|}{20}&\textbf{58.7$_{\uparrow\text{0.8}}$} & 31.7$_{\downarrow\text{3.6}}$ & 67.1$_{\downarrow\text{0.1}}$ & 43.3$_{\downarrow\text{0.2}}$ & 47.8$_{\downarrow\text{0.9}}$ & \multicolumn{1}{c|}{87.4$_{\text{0.0}}$} &56.9$_{\downarrow\text{1.0}}$ & 32.2$_{\downarrow\text{3.0}}$ & 67.1$_{\downarrow\text{0.1}}$ & 43.3$_{\downarrow\text{0.2}}$ & 48.1$_{\downarrow\text{0.7}}$ & 87.1$_{\downarrow\text{0.3}}$ \\ 
\multicolumn{1}{c|}{21}&57.7$_{\downarrow\text{0.2}}$ & 31.7$_{\downarrow\text{3.5}}$ & 66.8$_{\downarrow\text{0.4}}$ & 43.4$_{\downarrow\text{0.1}}$ & \textbf{49.2$_{\uparrow\text{0.4}}$} & \multicolumn{1}{c|}{\underline{87.0$_{\downarrow\text{0.4}}$}} &56.4$_{\downarrow\text{1.5}}$ & 32.9$_{\downarrow\text{2.3}}$ & 65.9$_{\downarrow\text{1.3}}$ & 43.5$_{\text{0.0}}$ & 47.5$_{\downarrow\text{1.3}}$ & 87.3$_{\downarrow\text{0.1}}$ \\ 
\multicolumn{1}{c|}{22}&58.0$_{\uparrow\text{0.1}}$ & 30.3$_{\downarrow\text{4.9}}$ & 67.1$_{\downarrow\text{0.1}}$ & 43.3$_{\downarrow\text{0.2}}$ & 48.5$_{\downarrow\text{0.3}}$ & \multicolumn{1}{c|}{87.8$_{\uparrow\text{0.3}}$} &55.1$_{\downarrow\text{2.8}}$ & \underline{30.1$_{\downarrow\text{5.1}}$} & \underline{64.9$_{\downarrow\text{2.3}}$} & 42.9$_{\downarrow\text{0.6}}$ & 47.2$_{\downarrow\text{1.6}}$ & 86.9$_{\downarrow\text{0.5}}$ \\ 
\multicolumn{1}{c|}{23}&\textbf{58.7$_{\uparrow\text{0.8}}$} & \underline{30.2$_{\downarrow\text{5.0}}$} & 67.6$_{\uparrow\text{0.4}}$ & \textbf{43.5$_{\text{0.0}}$} & 47.5$_{\downarrow\text{1.3}}$ & \multicolumn{1}{c|}{87.6$_{\uparrow\text{0.2}}$} &56.4$_{\downarrow\text{1.4}}$ & 30.7$_{\downarrow\text{4.5}}$ & 67.5$_{\uparrow\text{0.3}}$ & \textbf{43.6$_{\uparrow\text{0.1}}$} & 48.8$_{\text{0.0}}$ & 87.6$_{\uparrow\text{0.1}}$ \\ 
\bottomrule
\end{tabular}
}}
\dimen0=0.49\textwidth
\dimen2=\wd0
\ifdim\wd2>\dimen2 \dimen2=\wd2 \fi
\ExplSyntaxOn
\fp_set:Nn \l_tmpa_fp { \dim_to_fp:n { \dimen0 } / \dim_to_fp:n { \dimen2 } }
\dimen4=\fp_to_dim:n { \l_tmpa_fp * \dim_to_fp:n { \wd0 } }
\dimen6=\fp_to_dim:n { \l_tmpa_fp * \dim_to_fp:n { \wd2 } }
\edef\ijcnnTblScale{\fp_to_decimal:N \l_tmpa_fp}
\ExplSyntaxOff
% 用“内容真实宽度 + 固定中缝”的方式，把整体居中；避免出现整体左偏/右偏的观感。
\makebox[\textwidth][c]{%
    \begin{minipage}[b]{\the\dimen4}\centering
    \textbf{(a) MMStar (InternVL2-1B)}\\[-1pt]
    \scalebox{\ijcnnTblScale}{\copy0}
    \end{minipage}%
    \hspace{0.02\textwidth}%
    \begin{minipage}[b]{\the\dimen6}\centering
    \textbf{(b) Benchmarks (InternVL2-1B)}\\[-1pt]
    \scalebox{\ijcnnTblScale}{\copy2}
    \end{minipage}%
}
\endgroup
\end{table*}

\begin{table*}[!t]
\centering
\begingroup
\vspace{0.4ex}
\setlength{\belowcaptionskip}{0pt}
\setlength{\tabcolsep}{0.52mm}
\renewcommand{\arraystretch}{0.72}
\scriptsize
\caption{MMStar results of LayerNorm-Tuning, LoRA ($r=32$), and Partial-Tuning for each layer and all-layer tuning (ALL).}
\label{tab:main:mmstar_layer_peft}
\begin{tabular}{@{}ccccccccccccccclcccccc@{}}
\toprule
\multirow{3}{*}{Layer} & \multicolumn{7}{c}{Layernorm-Tuning} & \multicolumn{7}{c}{LoRA ($r=32$)} & \multicolumn{7}{c}{Partial-Tuning} \\ \cmidrule(l){2-22} 
 & OR & CP & FGP & IR & LR & MA & S\&T & OR & CP & FGP & IR & LR & MA & S\&T & \multicolumn{1}{c}{OR} & CP & FGP & IR & LR & MA & S\&T \\ \midrule

% \begin{equation}
% X_{\textit{scaled}} = \frac{X - X_{\textit{min}}}{X_{\textit{max}} - X_{\textit{min}}} \times \left(\text{Range}_{\textit{max}} - \text{Range}_{\textit{min}}\right)
% + \text{Range}_{\textit{min}}
% \end{equation}
% where $X$ is the original data, $X_{\textit{min}}$ and $X_{\textit{max}}$ represent the minimum and maximum values of each layers. The parameters $\text{Range}_{\textit{min}}$ and $\text{Range}_{\textit{max}}$ denote the minimum and maximum of the desired feature range.

\multicolumn{1}{c|}{0}&\multicolumn{1}{c|}{43.2} & \textbf{64.0} & 32.4 & \underline{50.8} & \underline{32.9} & \textbf{42.4} & \multicolumn{1}{c|}{36.4} &\multicolumn{1}{c|}{43.4} & 62.8 & 32.8 & 52.0 & 34.5 & 40.8 & \multicolumn{1}{c|}{37.6} &\multicolumn{1}{c|}{43.7} &\multicolumn{1}{c}{\textbf{64.0}} & 33.2 & 52.0 & 34.5 & 40.8 & 37.6 \\
\multicolumn{1}{c|}{1}&\multicolumn{1}{c|}{\textbf{43.8}} & 63.2 & 32.4 & 53.2 & 34.5 & \textbf{42.4} & \multicolumn{1}{c|}{36.8} &\multicolumn{1}{c|}{\underline{43.1}} & 62.8 & 32.8 & 51.6 & 34.5 & 41.2 & \multicolumn{1}{c|}{\underline{35.6}} &\multicolumn{1}{c|}{43.3} &\multicolumn{1}{c}{62.8} & 32.4 & 52.8 & \underline{32.9} & 41.2 & 37.6 \\
\multicolumn{1}{c|}{2}&\multicolumn{1}{c|}{43.6} & 63.2 & \textbf{34.4} & 52.0 & 34.1 & 40.8 & \multicolumn{1}{c|}{37.2} &\multicolumn{1}{c|}{43.7} & 62.4 & \underline{32.0} & 53.2 & \textbf{34.9} & 41.6 & \multicolumn{1}{c|}{38.0} &\multicolumn{1}{c|}{43.3} &\multicolumn{1}{c}{63.6} & 32.4 & 52.8 & 34.5 & 41.2 & \underline{35.2} \\
\multicolumn{1}{c|}{3}&\multicolumn{1}{c|}{43.6} & 63.2 & 33.6 & 51.6 & 34.5 & 42.0 & \multicolumn{1}{c|}{36.8} &\multicolumn{1}{c|}{43.4} & 63.6 & 32.8 & 51.6 & 34.1 & 42.0 & \multicolumn{1}{c|}{36.4} &\multicolumn{1}{c|}{43.6} &\multicolumn{1}{c}{\textbf{64.0}} & 31.2 & \textbf{53.6} & 34.5 & 41.2 & 37.2 \\
\multicolumn{1}{c|}{4}&\multicolumn{1}{c|}{43.6} & 63.2 & 33.2 & 52.8 & 34.1 & 41.2 & \multicolumn{1}{c|}{36.8} &\multicolumn{1}{c|}{43.7} & 62.4 & \textbf{35.2} & 52.8 & 34.1 & \underline{40.4} & \multicolumn{1}{c|}{37.2} &\multicolumn{1}{c|}{\underline{42.8}} &\multicolumn{1}{c}{\underline{62.0}} & 31.6 & 52.8 & 33.7 & 40.4 & 36.4 \\
\multicolumn{1}{c|}{5}&\multicolumn{1}{c|}{\textbf{43.8}} & \underline{62.0} & 32.4 & 52.8 & 34.5 & 42.0 & \multicolumn{1}{c|}{\textbf{38.8}} &\multicolumn{1}{c|}{43.9} & 63.6 & 34.0 & \textbf{53.6} & \underline{33.7} & 41.2 & \multicolumn{1}{c|}{37.2} &\multicolumn{1}{c|}{\textbf{43.8}} &\multicolumn{1}{c}{\textbf{64.0}} & 32.8 & 52.8 & 34.5 & 41.6 & 37.2 \\
\multicolumn{1}{c|}{6}&\multicolumn{1}{c|}{43.2} & 63.2 & 32.4 & 52.4 & 34.5 & 40.8 & \multicolumn{1}{c|}{\underline{35.6}} &\multicolumn{1}{c|}{43.6} & 63.6 & 34.4 & 52.0 & 34.1 & 41.2 & \multicolumn{1}{c|}{36.4} &\multicolumn{1}{c|}{43.7} &\multicolumn{1}{c}{63.2} & \textbf{34.0} & 52.8 & 33.7 & 40.4 & 38.0 \\
\multicolumn{1}{c|}{7}&\multicolumn{1}{c|}{43.4} & 63.6 & 32.0 & 52.0 & 34.5 & 41.2 & \multicolumn{1}{c|}{36.8} &\multicolumn{1}{c|}{43.8} & 62.8 & 33.2 & 52.0 & 34.5 & 42.0 & \multicolumn{1}{c|}{38.0} &\multicolumn{1}{c|}{43.4} &\multicolumn{1}{c}{63.6} & 33.6 & \underline{51.2} & 34.1 & 40.0 & 38.0 \\ \midrule
\multicolumn{1}{c|}{8}&\multicolumn{1}{c|}{43.4} & 63.6 & 32.8 & 53.2 & 33.7 & 41.6 & \multicolumn{1}{c|}{\underline{35.6}} &\multicolumn{1}{c|}{43.5} & 63.2 & 33.6 & 52.0 & 34.1 & 41.2 & \multicolumn{1}{c|}{36.8} &\multicolumn{1}{c|}{43.1} &\multicolumn{1}{c}{63.2} & 32.8 & 51.6 & 34.5 & 40.0 & 36.4 \\
\multicolumn{1}{c|}{9}&\multicolumn{1}{c|}{43.3} & 62.8 & 32.8 & 52.0 & 34.5 & 41.6 & \multicolumn{1}{c|}{36.0} &\multicolumn{1}{c|}{\textbf{44.1}} & \textbf{64.0} & 34.4 & 52.4 & 34.1 & 41.6 & \multicolumn{1}{c|}{38.0} &\multicolumn{1}{c|}{43.7} &\multicolumn{1}{c}{62.8} & 33.2 & 52.4 & 34.5 & 40.8 & \textbf{38.4} \\
\multicolumn{1}{c|}{10}&\multicolumn{1}{c|}{43.3} & 63.2 & 32.4 & 51.6 & 34.1 & 42.0 & \multicolumn{1}{c|}{36.4} &\multicolumn{1}{c|}{43.4} & 62.8 & 33.2 & 52.0 & \underline{33.7} & \underline{40.4} & \multicolumn{1}{c|}{38.0} &\multicolumn{1}{c|}{43.4} &\multicolumn{1}{c}{63.2} & 32.0 & 52.4 & 33.7 & 40.8 & 38.0 \\
\multicolumn{1}{c|}{11}&\multicolumn{1}{c|}{43.7} & 63.2 & 32.8 & \textbf{53.6} & 34.5 & 41.2 & \multicolumn{1}{c|}{36.8} &\multicolumn{1}{c|}{43.3} & 63.2 & 33.6 & 51.6 & 34.1 & 41.6 & \multicolumn{1}{c|}{\underline{35.6}} &\multicolumn{1}{c|}{43.5} &\multicolumn{1}{c}{63.2} & 32.8 & 52.8 & 35.3 & \underline{39.6} & 37.2 \\
\multicolumn{1}{c|}{12}&\multicolumn{1}{c|}{43.4} & 63.6 & 32.8 & 52.4 & 34.1 & 41.2 & \multicolumn{1}{c|}{36.4} &\multicolumn{1}{c|}{43.4} & 62.8 & 33.2 & 52.4 & 34.1 & \underline{40.4} & \multicolumn{1}{c|}{37.6} &\multicolumn{1}{c|}{43.6} &\multicolumn{1}{c}{63.6} & \textbf{34.0} & 52.0 & 34.1 & 40.4 & 37.6 \\
\multicolumn{1}{c|}{13}&\multicolumn{1}{c|}{43.2} & \underline{62.0} & 33.2 & 51.6 & 33.7 & 41.2 & \multicolumn{1}{c|}{37.6} &\multicolumn{1}{c|}{43.8} & 63.2 & 33.6 & \textbf{53.6} & 34.1 & 41.2 & \multicolumn{1}{c|}{36.8} &\multicolumn{1}{c|}{43.4} &\multicolumn{1}{c}{63.2} & 32.4 & 52.8 & 33.3 & 41.6 & 36.8 \\
\multicolumn{1}{c|}{14}&\multicolumn{1}{c|}{43.3} & 63.6 & 32.4 & 51.6 & 33.7 & 40.8 & \multicolumn{1}{c|}{37.6} &\multicolumn{1}{c|}{43.4} & 62.8 & 33.2 & 52.4 & 34.5 & 41.6 & \multicolumn{1}{c|}{36.0} &\multicolumn{1}{c|}{43.4} &\multicolumn{1}{c}{63.2} & 33.2 & 52.8 & 33.7 & 40.4 & 36.8 \\
\multicolumn{1}{c|}{15}&\multicolumn{1}{c|}{43.3} & 63.2 & 33.2 & 51.6 & 33.7 & 40.8 & \multicolumn{1}{c|}{37.2} &\multicolumn{1}{c|}{43.9} & 63.6 & 33.2 & 52.8 & \underline{33.7} & \textbf{42.4} & \multicolumn{1}{c|}{37.6} &\multicolumn{1}{c|}{\textbf{43.8}} &\multicolumn{1}{c}{62.8} & 33.2 & 52.4 & 34.9 & 41.6 & 37.6 \\ \midrule
\multicolumn{1}{c|}{16}&\multicolumn{1}{c|}{43.3} & 63.2 & 32.4 & 52.4 & \underline{32.9} & 42.0 & \multicolumn{1}{c|}{36.8} &\multicolumn{1}{c|}{43.6} & 63.6 & 33.6 & 52.4 & 34.1 & 40.8 & \multicolumn{1}{c|}{37.2} &\multicolumn{1}{c|}{43.5} &\multicolumn{1}{c}{63.6} & 33.2 & 52.0 & 34.1 & 41.6 & 36.4 \\
\multicolumn{1}{c|}{17}&\multicolumn{1}{c|}{\underline{42.9}} & 63.2 & 32.0 & 52.0 & \underline{32.9} & 40.8 & \multicolumn{1}{c|}{36.4} &\multicolumn{1}{c|}{43.6} & 63.6 & 33.6 & 51.6 & 34.1 & 41.6 & \multicolumn{1}{c|}{37.2} &\multicolumn{1}{c|}{42.9} &\multicolumn{1}{c}{62.8} & 32.0 & 51.6 & 33.3 & 40.8 & 36.8 \\
\multicolumn{1}{c|}{18}&\multicolumn{1}{c|}{43.7} & 63.2 & 32.8 & 52.4 & \textbf{34.9} & 41.2 & \multicolumn{1}{c|}{37.6} &\multicolumn{1}{c|}{43.8} & 63.6 & 32.8 & 52.8 & 34.5 & 40.8 & \multicolumn{1}{c|}{38.0} &\multicolumn{1}{c|}{42.9} &\multicolumn{1}{c}{62.8} & 32.0 & 52.0 & 34.1 & 41.2 & \underline{35.2} \\
\multicolumn{1}{c|}{19}&\multicolumn{1}{c|}{43.6} & \textbf{64.0} & 32.0 & 52.4 & 34.1 & 41.2 & \multicolumn{1}{c|}{37.6} &\multicolumn{1}{c|}{43.5} & 62.4 & 34.0 & \underline{51.2} & 34.1 & 42.0 & \multicolumn{1}{c|}{37.2} &\multicolumn{1}{c|}{43.6} &\multicolumn{1}{c}{63.6} & 32.4 & 53.2 & 34.1 & \textbf{42.0} & 36.0 \\
\multicolumn{1}{c|}{20}&\multicolumn{1}{c|}{43.4} & 63.2 & \underline{31.6} & 52.8 & 34.5 & \underline{40.4} & \multicolumn{1}{c|}{38.0} &\multicolumn{1}{c|}{43.4} & 63.2 & 32.8 & \underline{51.2} & \textbf{34.9} & 41.6 & \multicolumn{1}{c|}{36.8} &\multicolumn{1}{c|}{43.7} &\multicolumn{1}{c}{63.2} & 33.2 & 53.2 & 33.7 & 40.8 & 38.0 \\
\multicolumn{1}{c|}{21}&\multicolumn{1}{c|}{43.5} & 62.8 & 32.8 & 52.4 & 33.7 & 42.0 & \multicolumn{1}{c|}{37.2} &\multicolumn{1}{c|}{43.6} & 62.8 & 33.6 & \underline{51.2} & 34.1 & 41.2 & \multicolumn{1}{c|}{\textbf{38.8}} &\multicolumn{1}{c|}{43.4} &\multicolumn{1}{c}{62.8} & 33.2 & 52.0 & 34.9 & 40.8 & 36.4 \\
\multicolumn{1}{c|}{22}&\multicolumn{1}{c|}{43.4} & 63.2 & 32.8 & 53.2 & 33.7 & 40.8 & \multicolumn{1}{c|}{36.8} &\multicolumn{1}{c|}{43.6} & \textbf{64.0} & 33.2 & 51.6 & 34.5 & 41.2 & \multicolumn{1}{c|}{36.8} &\multicolumn{1}{c|}{43.5} &\multicolumn{1}{c}{63.6} & 32.4 & 52.0 & 33.7 & 41.6 & 37.6 \\
\multicolumn{1}{c|}{23}&\multicolumn{1}{c|}{43.6} & \underline{62.0} & 34.0 & 52.0 & 34.5 & 41.6 & \multicolumn{1}{c|}{37.2} &\multicolumn{1}{c|}{43.4} & \underline{62.0} & 32.8 & 52.0 & 34.1 & 42.0 & \multicolumn{1}{c|}{37.2} &\multicolumn{1}{c|}{43.4} &\multicolumn{1}{c}{63.6} & \underline{30.8} & 52.0 & 34.1 & 41.2 & \textbf{38.4} \\
\midrule
\rowcolor[HTML]{C0C0C0}
\multicolumn{1}{c|}{\cellcolor[HTML]{C0C0C0}ALL} & \multicolumn{1}{c|}{\cellcolor[HTML]{C0C0C0}43.2} & 63.6 & 32.8 & 52.4 & \underline{32.9} & 40.8 & \multicolumn{1}{c|}{36.4} & \multicolumn{1}{c|}{43.6} & 62.8 & 33.2 & \textbf{53.6} & 34.1 & 41.6 & \multicolumn{1}{c|}{36.0} & \multicolumn{1}{c|}{43.6} & 62.4 & 33.6 & 51.6 & \textbf{35.7} & 41.6 & 36.8  \\
%\multicolumn{1}{c|}{\cellcolor[HTML]{C0C0C0}ALL} & \multicolumn{1}{c|}{\cellcolor[HTML]{C0C0C0}29.1} & 27.6 & 24.0 & 29.6 & 21.6 & 40.0 & \multicolumn{1}{c|}{32.0} & \multicolumn{1}{c|}{42.5} & 62.8 & 31.6 & 50.0 &  30.4&42.8 & \multicolumn{1}{c|}{37.2} & \multicolumn{1}{c|}{41.8} & 56.4 & 29.6 & 50.8 &  34.4& 43.2 &36.4  \\ 
\bottomrule
\end{tabular}
\endgroup
\end{table*}

\subsection{Paired Significance Testing on Aggregated Benchmarks}
We conduct significance testing on the pooled six-benchmark set (ChartQA, HallusionBench, MMBench, OCRVQA, RealWorldQA, and ScienceQA) using a two-sided McNemar test and a paired percentile bootstrap 95\% CI (27k resamples) for the accuracy difference $\Delta$.

VeRA (layer~1) improves pooled accuracy by $\Delta=+2.14$pp (95\% CI $[1.63,\,2.64]$pp; $p=1.2\times 10^{-16}$), and LoRA (layer~2) by $\Delta=+2.08$pp (95\% CI $[1.57,\,2.58]$pp; $p=8.236\times 10^{-16}$), mainly driven by HallusionBench and RealWorldQA; OCRVQA slightly decreases for VeRA ($-1.23$pp; $p=0.1038$). Partial-Tuning (layer~13) also yields a small HallusionBench gain ($\Delta=+1.58$pp; $p=0.02007$).
In summary, probing-selected layers provide statistically significant gains on the pooled benchmarks, with improvements concentrated on HallusionBench and RealWorldQA.

Across both backbones and multiple PEFT families, perturbation-robust layers are distributed in both shallow and deep blocks and usually deliver stronger PEFT gains, supporting pre-PEFT probing as a practical layer-selection signal.

\section{Conclusion}
We study pre-PEFT layer selection for VLM vision encoders through Q/K/V weight statistics and perturbation probing. Across two backbones, layers that are more robust to perturbations tend to yield better PEFT outcomes, often together with larger weight norms and higher condition numbers in our evaluated settings. These results suggest that robustness-guided selection of a small subset of layers can reduce trainable parameters while remaining competitive with broader tuning in limited-data regimes.

\section*{Acknowledgment}
This work is partly supported by the Project granted by the Ministry of Agriculture and Rural Affairs, the Key Research and Development Program of Heilongjiang Province under Grant Nos. 2022ZX01A22, 2021ZXJ05A03, the National Natural Science Foundation of China  under Grant Nos. 62350710797, 61972114, 62106061, the National Science and Technology Major Project of China under Grant Nos. 2021ZD0110901, the collaborative innovation and promotion system of the modern agricultural industry technology for watermelon and melon in Heilongjiang Province.

% Bibliography: IEEEtran style, using only entries that are
% actually cited in the text (no nocite{*}, to avoid including
% template/example references from the .bib file)。
% 使用当前目录下的 ijcnn26.bib，
% 其包含正文与附录中实际用到的所有引用条目。
\bibliographystyle{IEEEtran}
\bibliography{ijcnn26}

@article{liu2022few,
  title={Few-shot parameter-efficient fine-tuning is better and cheaper than in-context learning},
  author={Liu, Haokun and Tam, Derek and Muqeeth, Mohammed and Mohta, Jay and Huang, Tenghao and Bansal, Mohit and Raffel, Colin A},
  journal={Advances in Neural Information Processing Systems},
  volume={35},
  pages={1950--1965},
  year={2022}
}

@inproceedings{hu2021lora,
  title={LoRA: Low-Rank Adaptation of Large Language Models},
  author={Hu, Edward J and Shen, Yelong and Wallis, Phillip and Allen-Zhu, Zeyuan and Li, Yuanzhi and Wang, Shean and Wang, Lu and Chen, Weizhu},
  booktitle={International Conference on Learning Representations},
  year={2022},
  url={https://openreview.net/forum?id=nZeVKeeFYf9}
}

@inproceedings{chen2024internvl,
  title={{InternVL}: Scaling Up Vision Foundation Models and Aligning for Generic Visual-Linguistic Tasks},
  author={Chen, Zhe and Wu, Jiannan and Wang, Wenhai and Su, Weijie and Chen, Guo and Xing, Sen and Zhong, Muyan and Zhang, Qinglong and Zhu, Xizhou and Lu, Lewei and others},
  booktitle={Proceedings of the IEEE/CVF Conference on Computer Vision and Pattern Recognition},
  pages={24185--24198},
  year={2024}
}

@misc{opengvlab2024internvl2,
  author={{OpenGVLab Team}},
  title={{InternVL2}},
  howpublished={Online},
  url={https://internvl.github.io/blog/2024-07-02-InternVL-2.0/},
  year={2024}
}

@article{bai2025qwen25vl,
  title={{Qwen2.5-VL} Technical Report},
  author={Bai, Shuai and Chen, Keqin and Liu, Xuejing and Wang, Jialin and Ge, Wenbin and Song, Sibo and Dang, Kai and Wang, Peng and Wang, Shijie and Tang, Jun and others},
  journal={arXiv preprint arXiv:2502.13923},
  year={2025}
}

@article{liu2024visual,
  title={Visual instruction tuning},
  author={Liu, Haotian and Li, Chunyuan and Wu, Qingyang and Lee, Yong Jae},
  journal={Advances in neural information processing systems},
  volume={36},
  year={2024}
}

@article{awadalla2023openflamingo,
  title={{OpenFlamingo}: An Open-Source Framework for Training Large Autoregressive Vision-Language Models},
  author={Awadalla, Anas and Gao, Irena and Gardner, Josh and Hessel, Jack and Hanafy, Yusuf and Zhu, Wanrong and Marathe, Kalyani and Bitton, Yonatan and Gadre, Samir and Sagawa, Shiori and others},
  journal={arXiv preprint arXiv:2308.01390},
  year={2023}
}

@inproceedings{houlsby2019parameter,
  title={Parameter-efficient transfer learning for NLP},
  author={Houlsby, Neil and Giurgiu, Andrei and Jastrzebski, Stanislaw and Morrone, Bruna and De Laroussilhe, Quentin and Gesmundo, Andrea and Attariyan, Mona and Gelly, Sylvain},
  booktitle={International conference on machine learning},
  pages={2790--2799},
  year={2019},
  organization={PMLR}
}

@inproceedings{radford2021learning,
  title={Learning Transferable Visual Models From Natural Language Supervision},
  author={Radford, Alec and Kim, Jong Wook and Hallacy, Chris and Ramesh, Aditya and Goh, Gabriel and Agarwal, Sandhini and Sastry, Girish and Askell, Amanda and Mishkin, Pamela and Clark, Jack and others},
  booktitle={International conference on machine learning},
  pages={8748--8763},
  year={2021},
  organization={PMLR}
}

@article{minaee2024large,
  title={Large language models: A survey},
  author={Minaee, Shervin and Mikolov, Tomas and Nikzad, Narjes and Chenaghlu, Meysam and Socher, Richard and Amatriain, Xavier and Gao, Jianfeng},
  journal={arXiv preprint arXiv:2402.06196},
  year={2024}
}

@article{yin2023survey,
  title={A survey on multimodal large language models},
  author={Yin, Shukang and Fu, Chaoyou and Zhao, Sirui and Li, Ke and Sun, Xing and Xu, Tong and Chen, Enhong},
  journal={arXiv preprint arXiv:2306.13549},
  year={2023}
}

@article{zhang2024vision,
  title={Vision-language models for vision tasks: A survey},
  author={Zhang, Jingyi and Huang, Jiaxing and Jin, Sheng and Lu, Shijian},
  journal={IEEE Transactions on Pattern Analysis and Machine Intelligence},
  year={2024},
  publisher={IEEE}
}

@inproceedings{jia2022visual,
  title={Visual Prompt Tuning},
  author={Jia, Menglin and Tang, Luming and Chen, Bor-Chun and Cardie, Claire and Belongie, Serge and Hariharan, Bharath and Lim, Ser-Nam},
  booktitle={European Conference on Computer Vision},
  pages={709--727},
  year={2022},
  organization={Springer}
}

@inproceedings{vaswani2017attention,
  title={Attention Is All You Need},
  author={Vaswani, Ashish and Shazeer, Noam and Parmar, Niki and Uszkoreit, Jakob and Jones, Llion and Gomez, Aidan N and Kaiser, {\L}ukasz and Polosukhin, Illia},
  booktitle={Advances in Neural Information Processing Systems},
  volume={30},
  pages={5998--6008},
  year={2017},
  url={https://papers.neurips.cc/paper/7181-attention-is-all-you-need}
}

@inproceedings{wang2021screen2words,
  title={Screen2words: Automatic mobile UI summarization with multimodal learning},
  author={Wang, Bryan and Li, Gang and Zhou, Xin and Chen, Zhourong and Grossman, Tovi and Li, Yang},
  booktitle={The 34th Annual ACM Symposium on User Interface Software and Technology},
  pages={498--510},
  year={2021}
}

@inproceedings{mishra2019ocr,
  title={{OCR-VQA}: Visual Question Answering by Reading Text in Images},
  author={Mishra, Anand and Shekhar, Shashank and Singh, Ajeet Kumar and Chakraborty, Anirban},
  booktitle={2019 international conference on document analysis and recognition (ICDAR)},
  pages={947--952},
  year={2019},
  organization={IEEE}
}

@article{kantharaj2022chart,
  title={Chart-to-Text: A Large-Scale Benchmark for Chart Summarization},
  author={Kantharaj, Shankar and Leong, Rixie Tiffany Ko and Lin, Xiang and Masry, Ahmed and Thakkar, Megh and Hoque, Enamul and Joty, Shafiq},
  journal={arXiv preprint arXiv:2203.06486},
  year={2022}
}

@article{lindstrom2022clevr,
  title={{CLEVR-Math}: A Dataset for Compositional Language, Visual and Mathematical Reasoning},
  author={Lindstr{\"o}m, Adam Dahlgren and Abraham, Savitha Sam},
  journal={arXiv preprint arXiv:2208.05358},
  year={2022}
}

@inproceedings{lu2022learn,
  title={Learn to Explain: Multimodal Reasoning via Thought Chains for Science Question Answering},
  author={Lu, Pan and Mishra, Swaroop and Xia, Tony and Qiu, Liang and Chang, Kai-Wei and Zhu, Song-Chun and Tafjord, Oyvind and Clark, Peter and Kalyan, Ashwin},
  booktitle={Advances in Neural Information Processing Systems},
  volume={35},
  year={2022},
  url={https://proceedings.neurips.cc/paper_files/paper/2022/hash/11332b6b6cf4485b84afadb1352d3a9a-Abstract-Conference.html}
}

@article{zhang2023llama,
  title={{LLaMA}-Adapter: Efficient Fine-Tuning of Language Models With Zero-Init Attention},
  author={Zhang, Renrui and Han, Jiaming and Liu, Chris and Gao, Peng and Zhou, Aojun and Hu, Xiangfei and Yan, Shilin and Lu, Pan and Li, Hongsheng and Qiao, Yu},
  journal={arXiv preprint arXiv:2303.16199},
  year={2023}
}

@article{masry2022chartqa,
  title={{ChartQA}: A Benchmark for Question Answering About Charts With Visual and Logical Reasoning},
  author={Masry, Ahmed and Long, Do Xuan and Tan, Jia Qing and Joty, Shafiq and Hoque, Enamul},
  journal={arXiv preprint arXiv:2203.10244},
  year={2022}
}

@inproceedings{guan2024hallusionbench,
  title={{HallusionBench}: An Advanced Diagnostic Suite for Entangled Language Hallucination and Visual Illusion in Large Vision-Language Models},
  author={Guan, Tianrui and Liu, Fuxiao and Wu, Xiyang and Xian, Ruiqi and Li, Zongxia and Liu, Xiaoyu and Wang, Xijun and Chen, Lichang and Huang, Furong and Yacoob, Yaser and others},
  booktitle={Proceedings of the IEEE/CVF Conference on Computer Vision and Pattern Recognition},
  pages={14375--14385},
  year={2024}
}

@article{liu2023mmbench,
  title={{MMBench}: Is Your Multi-Modal Model an All-Around Player?},
  author={Liu, Yuan and Duan, Haodong and Zhang, Yuanhan and Li, Bo and Zhang, Songyang and Zhao, Wangbo and Yuan, Yike and Wang, Jiaqi and He, Conghui and Liu, Ziwei and others},
  journal={arXiv preprint arXiv:2307.06281},
  year={2023}
}

@article{chen2024we,
        title={Are We on the Right Way for Evaluating Large Vision-Language Models?},
        author={Chen, Lin and Li, Jinsong and Dong, Xiaoyi and Zhang, Pan and Zang, Yuhang and Chen, Zehui and Duan, Haodong and Wang, Jiaqi and Qiao, Yu and Lin, Dahua and others},
        journal={arXiv preprint arXiv:2403.20330},
        year={2024}
}

@inproceedings{dosovitskiy2020image,
  title={An Image Is Worth 16x16 Words: Transformers for Image Recognition at Scale},
  author={Dosovitskiy, Alexey and Beyer, Lucas and Kolesnikov, Alexander and Weissenborn, Dirk and Zhai, Xiaohua and Unterthiner, Thomas and Dehghani, Mostafa and Minderer, Matthias and Heigold, Georg and Gelly, Sylvain and Uszkoreit, Jakob and Houlsby, Neil},
  booktitle={International Conference on Learning Representations},
  year={2021},
  url={https://openreview.net/forum?id=YicbFdNTTy}
}

@article{yang2024qwen2,
  title={{Qwen2} Technical Report},
  author={Yang, An and Yang, Baosong and Hui, Binyuan and Zheng, Bo and Yu, Bowen and Zhou, Chang and Li, Chengpeng and Li, Chengyuan and Liu, Dayiheng and Huang, Fei and others},
  journal={arXiv preprint arXiv:2407.10671},
  year={2024}
}

@article{kopiczko2023vera,
  title={{VeRA}: Vector-Based Random Matrix Adaptation},
  author={Kopiczko, Dawid Jan and Blankevoort, Tijmen and Asano, Yuki Markus},
  journal={arXiv preprint arXiv:2310.11454},
  year={2023}
}

@article{coppola2024investigating,
  title={Investigating and unmasking feature-level vulnerabilities of CNNs to adversarial perturbations},
  author={Coppola, Davide and Lee, Hwee Kuan},
  journal={arXiv preprint arXiv:2405.20672},
  year={2024}
}

@article{nokabadi2024reproducibility,
  title={Reproducibility Study on Adversarial Attacks Against Robust Transformer Trackers},
  author={Nokabadi, Fatemeh Nourilenjan and Lalonde, Jean-Fran{\c{c}}ois and Gagn{\'e}, Christian},
  journal={arXiv preprint arXiv:2406.01765},
  year={2024}
}

@inproceedings{he2019rethinking,
  title={Rethinking imagenet pre-training},
  author={He, Kaiming and Girshick, Ross and Doll{\'a}r, Piotr},
  booktitle={Proceedings of the IEEE/CVF international conference on computer vision},
  pages={4918--4927},
  year={2019}
}

@inproceedings{glorot2011deep,
  title={Deep sparse rectifier neural networks},
  author={Glorot, Xavier and Bordes, Antoine and Bengio, Yoshua},
  booktitle={Proceedings of the fourteenth international conference on artificial intelligence and statistics},
  pages={315--323},
  year={2011},
  organization={JMLR Workshop and Conference Proceedings}
}

@misc{xai_realworldqa_2024,
  author = {{xAI}},
  title = {{RealWorldQA}},
  year = {2024},
  howpublished = {Hugging Face},
  url = {https://huggingface.co/datasets/xai-org/RealworldQA}
}

\end{document}